\documentclass[journal]{IEEEtran}
\usepackage{afterpage}
\usepackage{subfigure}
\usepackage{subcaption}
\usepackage{hyperref}
\usepackage{adjustbox} 
\usepackage{rotating} 
\usepackage{subcaption} 
\usepackage{graphicx} 
\usepackage{amsmath}
\usepackage{amssymb}
\usepackage{comment}
\usepackage{graphicx} 
\usepackage{multirow} 
\usepackage{makecell} 
\usepackage{array} 
\usepackage{booktabs} 
\usepackage[table,xcdraw]{xcolor} 
\usepackage{booktabs}
\usepackage{xcolor}
\usepackage{tikz}
\usepackage[ruled,vlined]{algorithm2e}
\usepackage{orcidlink}
\definecolor{colComments}{rgb}{1,0,0}
\newcommand{\remind}[1]{{\color{colComments} \textbf{#1}  }}
\usepackage[normalem]{ulem} 

\ifCLASSINFOpdf
\else
\fi
\begin{document}
\newcommand{\prop}{FGump}

\newcommand{\labsty}[1]{{\footnotesize #1}}

\newcommand{\lablidarCHM}{\labsty{LiDAR-CHM}}
\newcommand{\lablidarDTM}{\labsty{LiDAR-DTM}}

\newcommand{\labCat}{\labsty{\prop}}
\newcommand{\labTSNN}{\labsty{TSNN}}
\newcommand{\labML}{\labsty{XGBoost}}
\newcommand{\labRF}{\labsty{RF}}
\newcommand{\labSKP}{\labsty{SKP}}
\newcommand{\labGLRT}{\labsty{GLRT}}
\newcommand{\labXGBoost}{\labsty{XGBoost}}
\newcommand{\labLightGBM}{\labsty{LightGBM}}
\newcommand{\labKNN}{\labsty{KNN}}

%
\title{

An Efficient Machine Learning Framework for Forest Height Estimation from Multi-Polarimetric Multi-Baseline SAR data
}
%
%
%

\author{Francesca Razzano \orcidlink{0009-0006-0755-0066},~\IEEEmembership{Student Member,~IEEE,}
        Wenyu Yang \orcidlink{0000-0002-5031-7212},~\IEEEmembership{Graduate Student Member,~IEEE,}\\
        Sergio Vitale \orcidlink{0000-0001-9784-0070},~\IEEEmembership{Member,~IEEE,} 
        Giampaolo Ferraioli \orcidlink{0000-0003-2441-0648},~\IEEEmembership{Senior Member,~IEEE,}\\        
        Silvia Liberata Ullo \orcidlink{0000-0001-6294-0581},~\IEEEmembership{Senior Member,~IEEE,}
        and~Gilda Schirinzi \orcidlink{0000-0002-9656-2969},~\IEEEmembership{Senior Member,~IEEE}
\thanks{F. Razzano, W. Yang, S. Vitale and G. Schirinzi are with Engineering Department, University of Naples Parthenope, Naples, Italy, email: [francesca.razzano002, wenyu.yang001]@studenti.uniparthenope.it, \nobreak and \nobreak [sergio.vitale, gilda.schirinzi]@uniparthenope.it}\nobreak 
\thanks{G. Ferraioli is with the Science and Technology Department, University of Naples Parthenope, Naples, Italy, email: giampaolo.ferraioli@uniparthenope.it} \nobreak 
\thanks{S. L. Ullo is with the Engineering Department, University of Sannio, Benevento, Italy, email: ullo@unisannio.it}}

%
%

\markboth{This paper has been submitted to IEEE TGRS. At the moment is under review}%
{Shell \MakeLowercase{\textit{et al.}}: An Efficient Machine Learning Framework for
Forest Height Estimation from Multi-Polarimetric
Multi-Baseline SAR data}

%



\maketitle
\begin{center}
   \color{blue}This paper has been submitted to IEEE TGRS. At the moment is under review.
\end{center}

\begin{abstract}
Accurate forest height estimation is crucial for climate change monitoring and carbon cycle assessment. Synthetic Aperture Radar (SAR), particularly in multi-channel configurations, has provided support for a long time in 3D forest structure reconstruction through model-based techniques. More recently, data-driven approaches using Machine Learning (ML) and Deep Learning (DL) have enabled new opportunities for forest parameter retrieval.
This paper introduces \prop{}, a forest height estimation framework by gradient boosting using multi-channel SAR processing with LiDAR profiles as Ground Truth (GT). Unlike typical ML and DL approaches that require large datasets and complex architectures, \prop{} ensures a strong balance between accuracy and computational efficiency, using a limited set of hand-designed features and avoiding heavy preprocessing (e.g., calibration and/or quantization).
Evaluated under both classification and regression paradigms, the proposed framework demonstrates that the regression formulation enables fine-grained, continuous estimations and avoids quantization artifacts by resulting in more precise measurements without rounding. Experimental results confirm that \prop{} outperforms State-of-the-Art (SOTA) AI-based and classical methods, achieving higher accuracy and significantly lower training and inference times, as demonstrated in our results.

\end{abstract}


%
\IEEEpeerreviewmaketitle

\section{Introduction}
\label{introduction}

\IEEEPARstart{B}{iomass} monitoring in forests is a critical component of sustainable environmental management and conservation efforts. 
One key aspect of biomass monitoring is the evaluation of tree vertical structure, which includes measuring tree height representing an important indicator of forest health, growth rates, and biomass distribution. 
Accurate monitoring of tree  height with high vertical resolution is crucial for assessing forest biomass, understanding ecological dynamics, and implementing sustainable forest practices \cite{8995799}, \cite{ENGLHART20111260}.
In situ forest monitoring, while providing the most accurate data parameters, is often costly and time-consuming. In contrast, Remote Sensing (RS) techniques offer significant advantages, including consistent, large-scale coverage and frequent revisit times, which help to overcome the challenges associated with traditional on-site techniques. RS methods that ensure underfoliage penetration and enable three-dimensional measurements, capturing both horizontal and vertical variability, are particularly well-suited for effective forest monitoring.

Synthetic Aperture Radar (SAR), due to its ability in penetrating forest structures, is a system able to provide information on the vertical distribution and internal composition of the area under investigation \cite{MB6}. 
Different SAR configurations and products, such as Polarimetric SAR (PolSAR), Polarimetric Interferometic SAR (PolInSAR) \cite{Cloude1998} \cite{Papathanassiou2001}, Tomographic SAR (TomoSAR) \cite{Frey2011} and Polarimetric Tomographic SAR (PolTomoSAR) \cite{Tebaldini2012}, have been extensively and effectively investigated in recent years to retrieve essential forest information. The common feature among these techniques is the use of multiple datasets (i.e., multi-channel data) related to the observed scene, creating constructive diversity: PolSAR utilizes different polarimetric data, PolInSAR employs polarimetric data from two different positions, TomoSAR exploits a stack of data acquired from different positions, and PolTomoSAR leverages polarimetric data from multiple positions \cite{8995799} \cite{Fornaro_book}.

In particular, TomoSAR is able to infer the reflectivity profiles along azimuth, range, and elevation co-ordinates by synthesizing  an additional aperture in the elevation direction \cite{MB6}. The three dimensional (3D) image is obtained by the coherent processing of multi-baseline (MB) data. The reconstruction accuracy of forest height depends significantly on how accurate the identification and positioning of the scattering phase centers from both the canopy and the ground are. PolTomoSAR techniques in the last decade have been widely employed to overcome the limitations of TomoSAR \cite{Tebaldini2012} \cite{ImSKP27} \cite{Guliaev2024}, thanks to the sensitivity of polirmetric information to shape, orientation and dielectric properties of scatterers, which facilitate the separation of ground and canopy contributions in order to improve the estimation accuracy of height profiles. Anyway, due to the great complexity of forested scenarios, it is extremely difficult to formulate accurate scattering models on which the inversion should be based, so that tomographic reconstructions are usually affected by inaccuracies and outliers.  
To overcome this problem, Machine Learning (ML) and Deep Learning (DL) approaches have been developed in the last years, showing a significant promise in improving the accuracy and efficiency of forest height estimation from a stack of multi-baseline polarimetric SAR images. Mainly, ML and DL based methods leverage the combined use of SAR and Light Detection and Ranging (LiDAR) data.

In the last years, different DL based approaches have been exploited. 
Beyond the definition of interesting architectures, as the case  of complex-valued convolutional neural networks (CV-CNN) \cite{8822431}, DL based forest height retrieval approaches are characterized mainly by the use of both multi-channel SAR (PolSAR, PolInSAR, TomoSAR and PolTomoSAR) data and LiDAR data.  \textit{Zhang et al.} in \cite{ZHANG2022123} presented a generative DL approach for forest height estimation, redefining the forest height inversion problem as a pan-sharpening process  between the low spatial resolution LiDAR profile, exhibiting a high height accuracy, and the high spatial resolution PolSAR and PolInSAR features.
In \cite{Yang2023} \textit{Yang et al.}, following the approach proposed in \cite{Budillon2019} of addressing the TomoSAR 3D height reconstruction as a classification problem, introduced TSNN, a Fully Connected Neural Network (FCNN) estimating the forest and ground heights using TomoSAR data as input and quantized LiDAR height values as reference class labels. The method proved to be very promising, but suffers from a high processing time and the effect of height profiles quantization.
\textit{Carcereri et al.} in \cite{10234548} investigated, instead, the potential of DL addressing forest height estimation as a regression problem, using TanDEM-X bistatic InSAR data and a fully convolutional neural network (CNN) framework. 
Although most of these methods employ airborne data, the use of spaceborne SAR and LiDAR data has also been investigated. In \cite{10049501}, the use of the Global Ecosystem Dynamics Investigation (GEDI) spaceborne LiDAR data and Tandem-X SAR images have been considered, highlighting the limitation due to the difficulties of penetration of X-band waveforms.
For this, \textit{Gupta et al.} \cite{GUPTA2022100817} incorporated a continuous coverage of multi-spectral optical and SAR data along with sparsely GEDI spaceborne LiDAR data in the ML models for mapping forests canopy height (Hcanopy) in the mixed tropical forests.
Recently, an alternative DL approach to perform the tomographic inversion has been presented in \cite{Berenger2023}, where a light-weight neural networks, trained with a single feed-forward pass using simulated data, is used for a fast reconstruction of the forest layer. Since the network training is based on synthetic data, a bias towards solutions complying with the very simple adopted model can be introduced, thus compromising the network generalization.

As for ML, the potential of different  methods such as Random Forest (RF), Rotation Forest, Canonical Correlation Forest (CCF), and Support Vector Machines (SVM), taking advantage of the combination of PoLSAR and LiDAR data for the forest height estimation, has been exploited in \cite{8469014}, \cite{8908748}, \cite{POURSHAMSI202179}. 
In \cite{8469014}, \cite{8908748}, SVM has been employed to extrapolate LiDAR-based canopy heights using PolInSAR inverted parameters, without assuming any specific data distribution. 
In \cite{POURSHAMSI202179} the potential of combining PolSAR data with a small subset of LiDAR data to achieve accurate height estimates across various forest height ranges, offering a scalable solution, is highlighted. 
Usually, the development of ML methods requires several crucial pre-processing steps aimed at the extraction of specifically-designed features.
For example, as shown in \cite{ZHANG2023103532},
for estimating forest height  from single-polarization
TomoSAR (SP-TomoSAR) and PolSAR data, the Capon estimator is employed to obtain the vertical power spectrum, while two polarization decomposition methods—H/A/alpha decomposition and Freeman-Durden decomposition—are used to extract the polarimetric features. The forest height estimated by the ML algorithms is compared with the Canopy Height Model 
(CHM) derived from LiDAR to verify accuracy.

While DL approaches usually provide accurate results by the solely exploitation of data, without the need of hand-designing specific features, the construction of a huge dataset and of deep and complex architecture are mandatory steps for achieving good performance. 
To the other hand, ML approaches can achieve interesting performance with limited data with a reduced complexity and higher computational efficiency at the cost of a precise design of input features \cite{Goodfellow2016}. 

On the basis of these considerations, although DL methods
perform outstandingly well in some application areas where large labeled datasets are available, such as computer vision and natural language understanding, for other types of datasets it is worthwhile to investigate  whether or not neural networks outperform other competing 
ML techniques. Therefore, an analysis on the application of most promising ML methods to SAR tomography can open new insights and premises for the development of new, more performant processing methods.

In this paper, \textit{\textbf{F}orest height estimation by \textbf{G}radient boosting \textbf{u}sing \textbf{m}ulti-channel SAR \textbf{p}rocessing} (\prop{}), a ML approach that can balance efficiency and accuracy with a limited number of input features, is proposed for the 3D forest height reconstruction. 



This framework introduces a novel, regression-driven interpretation of SAR-based retrieval of canopy and ground height profiles, capable of operating without the need for heavy pre-processing (e.g., calibration and/or quantization), while maintaining strong generalization. These characteristics make it highly suitable for large-scale, time-constrained forestry applications and pave the way for operational integration of ML-based height mapping systems using SAR data. 

This work highlights the effectiveness of a new framework using CatBoost, an ensemble learning technique that has achieved State-of-the-Art (SOTA) performance in diverse ML applications by combining the predictions from two or more base models \cite{Mienye2022}. It is shown that the prsented method outperforms conventional approaches while offering key advantages such as implementation simplicity and significantly reduced training time, features that make it especially well-suited for forest height estimation in complex vegetative environments.
The main contribution of this framework are outlined below:
\begin{itemize}

    \item \textbf{Superior computational efficiency and accuracy}: The proposed framework demonstrates substantially lower training and inference times compared to both DL and other ML methods achieving better accuracy. This efficiency makes the model particularly suitable for time-sensitive or iterative applications and paves the way for integrating rapid modeling strategies in future work;

    \item \textbf{Performant regression paradigm enabling fine-grained estimation}: By adopting a regression-based formulation, the proposed method avoids discretization-related artifacts and preserves the continuity of the canopy and terrain height information. As a result, the quantization step typically required in classification-based tasks using LiDAR reference data is eliminated, further streamlining the workflow;
    
    \item \textbf{Robustness}: A comprehensive experimental analysis demonstrates the effectiveness of the proposed solution under various pre-processing hyperparameter configurations including different spatial window sizes for covariance matrix estimation, SAR calibration settings, and learning paradigms offering valuable insights into the model’s robustness and sensitivity to diverse design choices.
    

\end{itemize}

In summary, the proposed framework demonstrates that accurate forest height estimation can be achieved using a regression-based ML approach that relies on compact, physically meaningful SAR features, without the need for extensive pre-processing. Its strong performance and low computational cost make it a promising solution for scalable and operational forest monitoring applications.
The structure of this work is as follows. Section \ref{introduction} provides an overview of the context and motivation of the study. Section \ref{methodology} presents the methodological framework: Subsection \ref{subs: costr_data} outlines the process of dataset construction, Subsection \ref{subs: theorycat} details the theoretical foundations of the CatBoost algorithm used in the proposed framework, and Subsection \ref{subs: set_cat} describes the training setup. Section \ref{results} presents the study area and dataset in Subsection \ref{subs: studyarea}, followed by a focused evaluation of the CatBoost model in Subsection \ref{subs: catb}, where the robustness of the proposed approach is assessed with respect to key aspects. A comparison with other State-of-the-Art (SOTA) ML algorithms is provided in Subsection \ref{subs: comp_ML}, alongside a discussion of the results in Subsection \ref{subs: analysis_result} and an analysis of computational efficiency in Subsection \ref{subs: compcost}. Lastly, Section \ref{sec: conclusion} synthesizes the key findings, discusses their implications, and suggests potential avenues for future research.

\section{Methodology}
\label{methodology}
 
In this section the novel FGump framework, as depicted in Fig. \ref{fig: model}, is described. First, the dataset construction process is introduced in subsection \ref{subs: costr_data}, then the considered ML algorithm is outlined in subsection \ref{subs: theorycat}, and finally details on its implementation, including the training settings used in the experiments, are  presented in subsection \ref{subs: set_cat}.


\subsection{Dataset Construction} \label{subs: costr_data}
The proposed methodology relies on a supervised ML method; therefore, a labeled training dataset is required. 

The dataset construction follows the same workflow proposed  by the same authors in \cite{Yang2023} where  a DL advancement on multi-channel SAR processing for height inversion have been recently proposed.
In particular, the proposed method relies on the use of both multi-polarimetric multi-baseline (MPMB) data and LiDAR data for the training step. 
For each fixed range-azimuth pixel in the MPMB image stack, $3N_b$ data values are available, with $N_b$ the number of acquisitions for each of the 3 polarizations HH, HV and VV. 

The polarimetric multi-baseline sample covariance matrix $\mathbf{R}$ of size $3N_b\times3N_b$ \cite{Fornaro_book} can be computed by means of a spatial averaging operation on a window surrounding the given pixel, thus obtaining:

\begin{equation}
\label{matrix}
\mathbf{R} = \begin{bmatrix}
\mathbf{C}_{11} & \mathbf{\Omega}_{12} & \cdots & \mathbf{\Omega}_{1N_b} \\
\mathbf{\Omega}_{21} & \mathbf{C}_{22} & \cdots & \mathbf{\Omega}_{2N_b} \\
\vdots & \vdots & \ddots & \vdots \\
\mathbf{\Omega}_{N_b1} & \mathbf{\Omega}_{N_b2} & \cdots & \mathbf{C}_{N_bN_b}
\end{bmatrix}.
\end{equation}
where the blocks $\mathbf{C}_{ii}$ and $\mathbf{\Omega}_{ik}$, with $i,k=1, \cdots , N_b$, are the $3\times3$ polarimetric correlation and mutual correlation matrices for each one of the $N_b$ acquisition trajectories.

Then, the $3N_b$ (real valued) diagonal elements, along with the $3N_b-1$ (complex valued) elements of the first row of the matrix  \eqref{matrix}, encapsulating both polarimetric and interferometric information, are extracted and used as inputs to the ML algorithm. 

Specifically, the complex valued elements of the first row are separated into real and imaginary parts, with the imaginary unit discarded. This process results in a real-valued input feature vector $\boldsymbol{x}$ for each range-azimuth pixel, sized $M \times 1$ where $M = 3N_b + 2 (3N_b - 1)$. 
The LiDAR-based height values are considered Ground Truth (GT) in this context. To align with the spatial resolution of SAR data, a spatial averaging operation is applied to the LiDAR-based CHM and Digital Terrain Model (DTM). The size of the averaging window matches the one used in the computation of the MPMB covariance matrix \eqref{matrix}, ensuring consistency in spatial sampling and facilitating direct comparison and integration of SAR and LiDAR-derived information for height estimation.

\subsection{ML algorithm: a gradient boosting approach} \label{subs: theorycat}
 Due to the aim of this work being the design of ML-based framework balancing efficiency and accuracy, the choice falls within the gradient boosting algorithms. In particular, the CatBoost algorithm has been considered.
 CatBoost (short for "Categorical Boosting") \cite{prokhorenkova2019catboostunbiasedboostingcategorical} is a gradient boosting library handling categorical features and outperforming previous implementations of gradient boosting in terms of accuracy on several popular datasets.
 
 Gradient boosting is a robust ML technique known for delivering SOTA results across a wide array of practical applications \cite{ROE2005577}. It has long been the go-to method for tackling learning problems involving heterogeneous features, noisy data, and complex dependencies, such as web search, recommendation systems, and weather forecasting. Essentially, gradient boosting constructs an ensemble predictor by applying gradient descent in a functional space. Theoretical foundations support this approach, demonstrating how strong predictors can be developed by iteratively and greedily combining weaker models (base predictors) \cite{dorogush2018catboostgradientboostingcategorical}. CatBoost introduces two pivotal algorithmic innovations: ordered boosting and a novel method for processing categorical features. Ordered boosting, a permutation-driven variant of the traditional gradient boosting algorithm, effectively prevents target leakage. Additionally, the new algorithm for handling categorical features enhances model performance. These innovations are integrated into CatBoost, an open-source library that surpasses current SOTA implementations of gradient boosted decision trees, such as Extreme Gradient Boosting (XGBoost) and Light Gradient-Boosting Machine (LightGBM), across a wide range of popular ML tasks.
 
Gradient boosting is a powerful ensemble technique in ML, primarily utilized to enhance the performance of predictive models by combining multiple weak learners. The fundamental principle behind gradient boosting is to construct the final predictive model incrementally by iteratively adding new models that correct the errors of the existing ensemble. 

Mathematically, a gradient boosting model can be expressed as:
\begin{equation}
F(\boldsymbol{x}) = \sum_{z=1}^{Z} h_z(\boldsymbol{x}),
\end{equation}
where:
\begin{itemize}
\item $\boldsymbol{x}$ is the input data vector;
    \item $F(\boldsymbol{x})$ is the final predictive model;
    \item $h_z(\boldsymbol{x})$ represents the individual weak learners, which are typically decision trees;
    \item $Z$ is the total number of weak learners in the ensemble.
\end{itemize}

The core idea of gradient boosting is to sequentially build each new model $h_z(\boldsymbol{x})$ in such a way that it minimizes the residual error between the target reference value and the current ensemble model $F_{z-1}(\boldsymbol{x})$. Each subsequent model focuses on the errors made by the combined models up to that point, effectively improving the overall prediction accuracy \cite{prokhorenkova2019catboostunbiasedboostingcategorical}.

To minimize the residual errors, each weak learner $h_z(\boldsymbol{x})$ is trained using a loss function $\mathcal{L}(y, F_{z-1}(\boldsymbol{x}) + h_z(\boldsymbol{x}))$, where:
\begin{itemize}
    \item $y$ represents the actual target values (the LiDAR reference in our framework);
    \item $F_{z-1}(\boldsymbol{x})$ is the current ensemble model without the $z$-th tree;
    \item the loss function $\mathcal{L}$ quantifies the discrepancy between the target values $y$ and the predictions $F_{z-1}(\boldsymbol{x}) + h_z(\boldsymbol{x})$.
\end{itemize}

The process can be described as follows:
\begin{enumerate}
    \item \textbf{Initialize the Model:} Starting with an initial model $F_0(\boldsymbol{x})$, which could be a simple constant value model, often chosen as the mean of the target values.
    \item \textbf{Iterative Improvement:} For each iteration $z$ from 1 to $Z$:
    \begin{enumerate}
        \item For each training input $\boldsymbol{x}_i$, compute the residual errors based on the current model $F_{z-1}(\boldsymbol{x}_i)$:
        \begin{equation}
        l_{i}^{(z)} = y_i - F_{z-1}(\boldsymbol{x}_i),
        \end{equation}
        where $l_{i}^{(z)}$ denotes the residual error for the $i$-th training instance at the $z$-th iteration.
        \item Train a new weak learner $h_z(\boldsymbol{x})$ to predict these residuals. This weak learner is essentially trying to approximate the negative gradient of the loss function:
        \begin{equation}
        h_z(\boldsymbol{x}) \approx -\nabla_{F_{z-1}(\boldsymbol{x})} \mathcal{L}(y, F_{z-1}(\boldsymbol{x})).
        \end{equation}
        \item Update the ensemble model by adding the new weak learner:
        \begin{equation}
        F_z(\boldsymbol{x}) = F_{z-1}(\boldsymbol{x}) + \nu h_z(\boldsymbol{x}),
        \end{equation}
        where $\nu$ is a learning rate parameter that controls the contribution of each weak learner.
    \end{enumerate}
\end{enumerate}

Gradient boosting can be viewed as performing gradient descent in a functional space. In this context, the ensemble model $F(\boldsymbol{x})$ is optimized by iteratively moving in the direction of the steepest descent, as defined by the negative gradient of the loss function. Each weak learner $h_z(\boldsymbol{x})$ approximates this negative gradient, thereby improving the predictive accuracy of the model step by step. The strength of gradient boosting lies in its solid theoretical foundation. By iteratively combining weaker models in a greedy manner, gradient boosting leverages the principle that a collection of weak learners can collectively form a strong learner. This approach is backed by the consistency of gradient descent, ensuring that each step incrementally reduces the overall prediction error.
Another notable strength of the CatBoost algorithm lies in its innovative method for computing leaf values during tree construction. Unlike traditional (Gradient Boosted Decision Tree) GBDT implementations that rely on heuristic-based or approximate solutions, CatBoost employs exact greedy algorithms and symmetric trees with oblivious splits, which contribute to improved generalization and effective overfitting control \cite{prokhorenkova2019catboostunbiasedboostingcategorical}. This strategy enables a more stable and regular tree structure, which is particularly advantageous in scenarios with high-cardinality categorical features or noisy data.
Moreover, CatBoost supports both CPU and GPU execution backends, leveraging optimized data structures and efficient memory usage. The GPU implementation is designed to parallelize histogram computations and tree building, leading to significantly faster training times compared to leading libraries such as XGBoost and LightGBM, especially for large-scale datasets \cite{dorogush2018catboostgradientboostingcategorical}.
In addition to training efficiency, CatBoost features a highly optimized CPU inference engine, which outperforms existing open-source implementations in prediction speed on ensembles of comparable depth and size. These advantages make CatBoost a compelling solution for both development and deployment in real-world applications.

\begin{figure*}[!t]
    \centering
    \includegraphics[width=18.5cm]{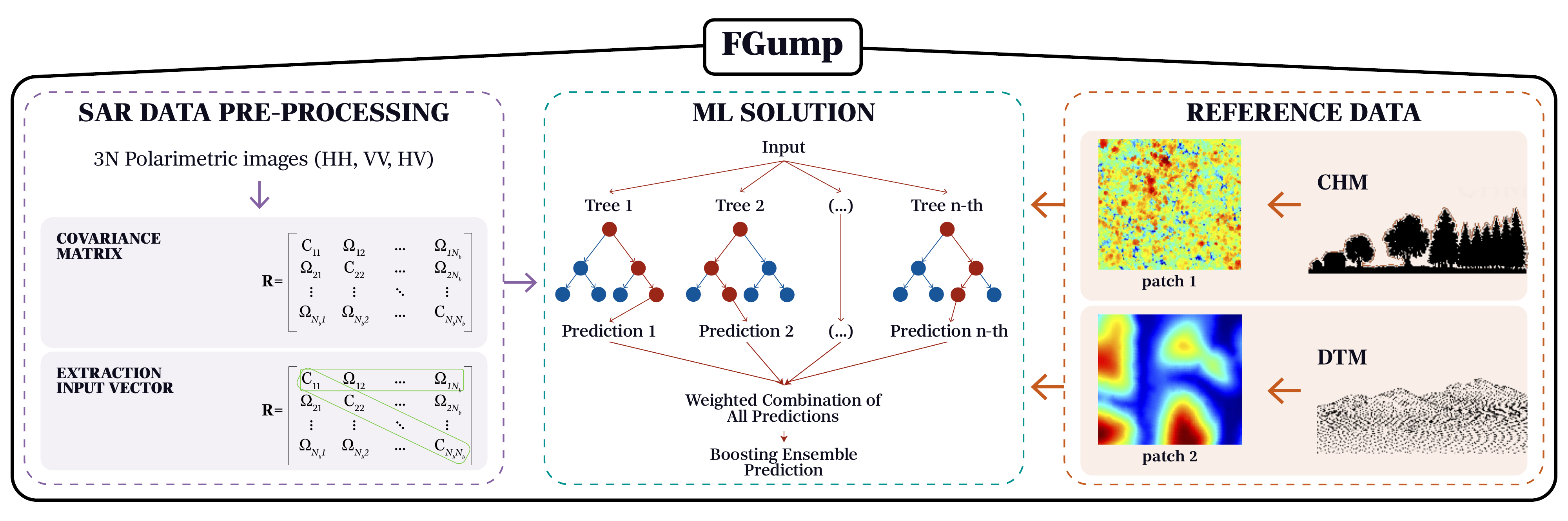}
    \caption{{FGump framework}: for each range-azimuth pixel of the MPMB image stack, the $3N_b$ diagonal elements of matrix $\mathbf{R}$ and the elements in the first row were extracted and used as inputs. The complex elements were separated into real and imaginary components before stacking. The reference labels were derived from LiDAR height values: a CHM for predicting forest height and a DTM for predicting ground height. The framework also includes the solution block based on a ML approach, which provides the prediction of values for both the CHM and DTM task}
    \label{fig: model}
\end{figure*}






\subsection{Algorithm Training} \label{subs: set_cat}
The proposed FGump framework has been implemented for addressing the height reconstruction from MPMB SAR data using the LiDAR data as reference, following two perspectives: classification and regression tasks.
A general workflow of the method is represented in Fig. \ref{fig: model}.

While the input data are represented by the 3$N_b$ diagonal elements and the elements in the first row of the matrix $\mathbf{R}$ for both classification and regression tasks, they basically differ for the considered GT and cost functions.

In the classification problem the reference labels were derived from the quantized LiDAR height values acting as classes to be predicted: a CHM was used for training aimed at predicting forest height, while a DTM was employed for predicting ground height. 
In this case, a weighted cross entropy for addressing the class-imbalance issue as in the following equation:

\begin{equation}
\mathcal{L}_c{} = 
\frac{
\sum_{i=1}^{N} w_i \log \left( 
\frac{e^{a_{i t_i}}}{\sum_{j=0}^{T-1} e^{a_{ij}}} 
\right)
}{
\sum_{i=1}^{N} w_i
}, \quad t_i \in \{0, \ldots, T-1\}
\end{equation}
where \(N\) is the number of training samples, \(T\) is the total number of classes, \(a_{ij}\) denotes the logit (i.e. the raw, unnormalized output generated by the model before applying any activation function) associated with the \(j\)-th class for the \(i\)-th sample, \(t_i\) is the true class label of the \(i\)-th sample, and \(w_i\) is a weight factor typically computed inversely to the frequency of the corresponding class. This formulation corresponds to the loss used by CatBoost when configured with loss function as MultiClass, optionally combined with class or sample weighting strategies.

In the regression problem, instead, the labels were represented by the non-quantized LiDAR data.
This approach preserves the full resolution of the original canopy or terrain height data, allowing the model to predict real-valued outputs and maintain fidelity to the true structural variability captured by LiDAR. Unlike classification, where height values are discretized into class intervals, regression maintains the continuous nature of the prediction target, which is particularly useful for precise height estimation tasks.

To train the regression model, a weighted Root Mean Squared Error (RMSE) loss function was adopted. This loss, which corresponds to the L2 norm of the prediction errors normalized by the sum of sample weights, penalizes larger errors more severely due to the squaring operation. It is defined as:

\begin{equation}
\label{L_r}
\mathcal{L}_r = \sqrt{\frac{\sum_{i=1}^{N} (F_Z(\boldsymbol{x}_i) - t_i)^2 w_i}{\sum_{i=1}^{N} w_i}}
\end{equation}
where \(N\) is the total number of samples, \(F_Z(\boldsymbol{x}_i)\) denotes the predicted value for the \(i\)-th input sample, \(y_i\) is the corresponding GT value, and \(w_i\) is a weight assigned to sample \(i\), which can be used to handle non-constant error variance in the data or to emphasize specific data subsets. The RMSE provides a scale-sensitive evaluation of the model’s predictive performance and is the default regression loss in several gradient boosting frameworks, including CatBoost when configured with loss function as RMSE.


\section{EXPERIMENTS}
\label{results}

This section presents the experimental analysis conducted to assess the performance of the proposed approach. The study area and the characteristics of the dataset are described in Subsection~\ref{subs: studyarea}. An in-depth evaluation of the proposed algorithm is then provided in Subsection~\ref{subs: catb}, focusing on its effectiveness in forest height estimation. Subsection~\ref{subs: comp_ML} presents a comparative analysis with other SOTA ML methods. The results are discussed in Subsection~\ref{subs: analysis_result}, while the computational efficiency of each method is analyzed in Subsection~\ref{subs: compcost}.

\subsection{Study Area and Dataset}
\label{subs: studyarea}
\begin{figure}[!h]
    \centering
    \includegraphics[width=\linewidth]{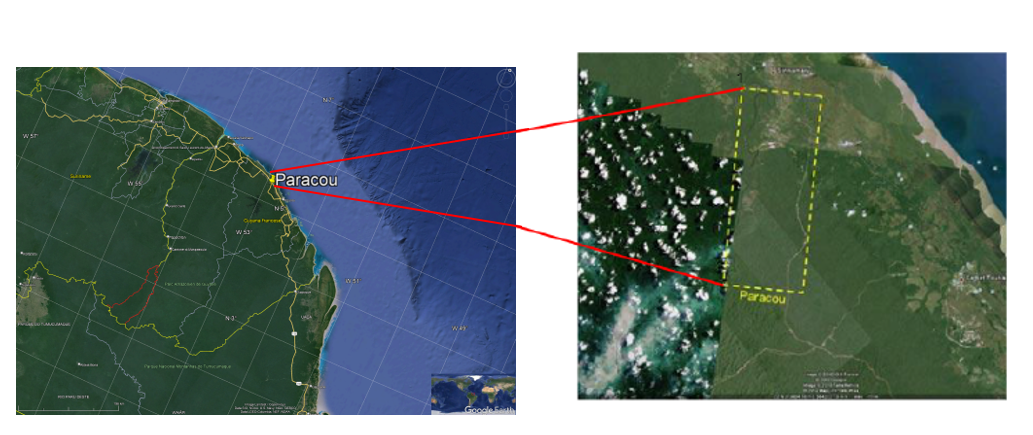}
    \caption{Paracou region of French Guiana area}
    \label{fig:area}
\end{figure}

The Paracou region of French Guiana, depicted in Fig. \ref{fig:area} is considered a case study. The site features diverse vegetation with various tree species, with vegetation heights ranging from 0 to 60 meters, and a relatively flat terrain with elevations between 0 and 40 meters. This area is representative of forest height and underlying topography reconstruction. 

\begin{figure*}[!t]
    \centering
    \includegraphics[width=\linewidth]{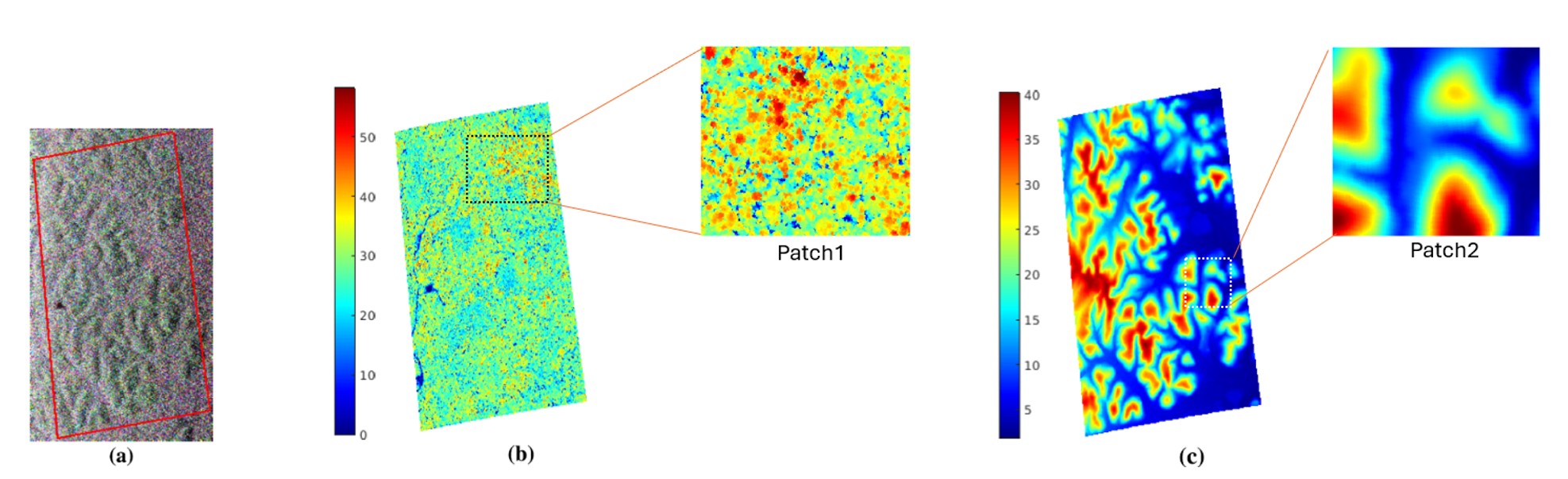}
    \caption{SAR data and LiDAR-derived data. (a) Pauli RGB image of the master acquisition. (b) LiDAR-derived CHM. (c) LiDAR-derived DTM of ROI in the Paracou area.}
    \label{fig:RGB_CHM_DTM}
\end{figure*}

The data stack consists of six fully polarimetric P-band SAR images, acquired by the ONERA SETHI airborne system on August 24, 2009, during the ESA’s TROPISAR campaign. The flight lines were spaceed out vertically, providing a Fourier vertical resolution of approximately 20 meters. SAR acquisition parameters are detailed in Table \ref{table:flight_parameters}.
The region of interest (ROI) is shown in Fig. \ref{fig:RGB_CHM_DTM}. In particular the Pauli RGB image of the master acquisition (red: HH-VV, green: 2 HV, and blue: HH + VV), the LiDAR CHM and the LiDAR DTM, are shown in the subfigures (a), (b) and (c) respectively.

\begin{table}[!ht]
    \centering
    \caption{Dataset description: Acquisition Parameter} \label{table:flight_parameters}
    \begin{tabular}{p{4cm} p{4cm}} 
        \toprule
        \rowcolor[HTML]{E0E0E0}
        \textbf{System Parameters} & \textbf{} \\
        \midrule
        Wavelength [m]& 0.7542\\
        Flight height [m]&  3962\\
        Incidence angle [deg] &  35.061\\
        Range resolution [m]  & 1 \\
        Azimuth resolution [m] & 1.245\\
        Polarization & Full-Pol\\
        Baselines [m] &  [0 -14 -30 -44 -60 -75]\\
        \bottomrule
    \end{tabular}
\end{table}

LiDAR-based data for the site are provided by the French Agricultural Research Center for International Development and the Guyafor project. 
Further details on LiDAR data processing are available in \cite{VINCENT201223}. LiDAR-based data are used as GT to train and verify the accuracy of the reconstructed results.
In particular, a testing patch of 280$\times$280 pixels from the ROI in the Paracou region was selected as the testing dataset for CHM estimation (see Patch1 in Fig. \ref{fig:RGB_CHM_DTM}). The remaining data were randomly divided into a training dataset (80\%) and a validating dataset (20\%). Similarly, a testing patch of 300$\times$300 pixels (see Patch2 in Fig. \ref{fig:RGB_CHM_DTM}) was chosen as the testing dataset for DTM estimation.
The remaining data were randomly divided into a training dataset (80\%) and a validating dataset (20\%). 

The number of baselines in TROPISAR is $N_b=6$, resulting in a covariance matrix of Eq. (\ref{matrix}) of size 18$\times$18. Thus, the input feature vector size for each range-azimuth pixel is 52$\times$1. For the analysis conducted in this manuscript, two datasets were constructed, distinguishing between phase calibrated (C) and non-calibrated (NC) SAR input data \cite{Aghababaee_PhaseCal}.

\subsection{Robustness Analysis for optimal configuration} \label{subs: catb}

 In order to perform the height estimation of the canopy and ground, some pre-processing steps of the input data are required. In this section, the robustness of the proposal with respect to some crucial aspects is investigated, in order to define the optimal configuration of the proposed algorithm. 
Specifically, we investigate the influence of three key factors:
\textit{(i)} the size of the spatial averaging window used in the covariance matrix estimation,
\textit{(ii)} the choice between regression and classification learning paradigms, and
\textit{(iii)} the impact of input data phase calibration \cite{Aghababaee_PhaseCal}.

Each aspect is systematically evaluated to understand its contribution to model performance, enabling a deeper insight into the behavior and robustness of proposal under different training conditions. This multidimensional analysis is crucial for optimizing the predictive capabilities of the model and tailoring it to the unique characteristics of forest monitoring based on SAR.
The RMSE between the LiDAR reference data and the predicted DTM and CHM values, obtained using different window sizes, is reported in Tables~\ref{tab: num catboost dtm} and~\ref{tab: num catboost chm}, respectively. Results are shown for both the classification and regression variants of proposed method, using calibrated and non-calibrated input data.


\begin{table}[h]
 \setlength{\tabcolsep}{4pt} 
\centering
\caption{RMSE values (expressed in meters) between predicted and reference DTM for different window sizes, using both classification and regression variants of CatBoost. Results are reported for C and NC input data.}
\begin{tabular}{lcccccc}
\toprule
 & \multicolumn{6}{c}{\textbf{Windows size}} \\
\cmidrule(lr){2-7}
 & 27$\times$27 & 31$\times$31 & 37$\times$37 & 41$\times$41 & 45$\times$45 & 49$\times$49 \\
\midrule
 Classification-NC & 2.62 & 2.40 & 2.27 & 2.15 & 2.08 & \textbf{1.96} \\
\rowcolor{gray!20} Classification-C & 2.64 & 2.38 & 2.25 & 2.14 & \textbf{2.01} & 2.012 \\
\midrule
 Regression-NC & 2.55 & 2.41 & 2.22 & 2.15 & 2.04 & \textbf{1.96} \\
\rowcolor{gray!20} Regression-C & 2.54 & 2.38 & 2.16 & 2.01 & 1.95 & \textbf{1.82} \\
\bottomrule
\end{tabular}
\label{tab: num catboost dtm}
\end{table}

\begin{table}
\setlength{\tabcolsep}{4pt}
\centering
\caption{RMSE values (expressed in meters) between predicted and reference CHM for different window sizes, using both classification and regression variants of proposed method. Results are reported for C and NC input data.
}
\begin{tabular}{lcccccc}
\toprule
 & \multicolumn{6}{c}{\textbf{Windows size}}  \\
\cmidrule(lr){2-7}
 & 27$\times$27 & 31$\times$31 & 37$\times$37 & 41$\times$41 & 45$\times$45 & 49$\times$49 \\
\midrule
Classification-NC & 2.83 & 2.60 & 2.36 & 2.22 & 2.13 & \textbf{2.12} \\
\rowcolor{gray!20} Classification-C & 2.79 & 2.57 & 2.35 & 2.20 & 2.13 & \textbf{2.02} \\
\midrule
 Regression-NC & 2.52 & 2.28 & 2.06 & 1.92 & 1.86 & \textbf{1.83} \\
\rowcolor{gray!20} Regression-C & 2.54 & 2.29 & 2.04 & 1.98 & 1.93 & \textbf{1.88} \\
\bottomrule
\end{tabular}
\label{tab: num catboost chm}

\end{table}

\begin{figure*}[ht]
    \centering
    
    \begin{tabular}{>{\centering\arraybackslash}m{0cm} c c c c c c}
        &  27$\times$27 &  31$\times$31 & 37$\times$37 &  41$\times$41 &  45$\times$45 &  49$\times$49 \\
         
        
        \multirow{4}{*}[5.4em]{\makecell{\hspace{-1mm}\rotatebox{90}{\footnotesize Classification-NC}}} & 
        \includegraphics[width=2.5cm]{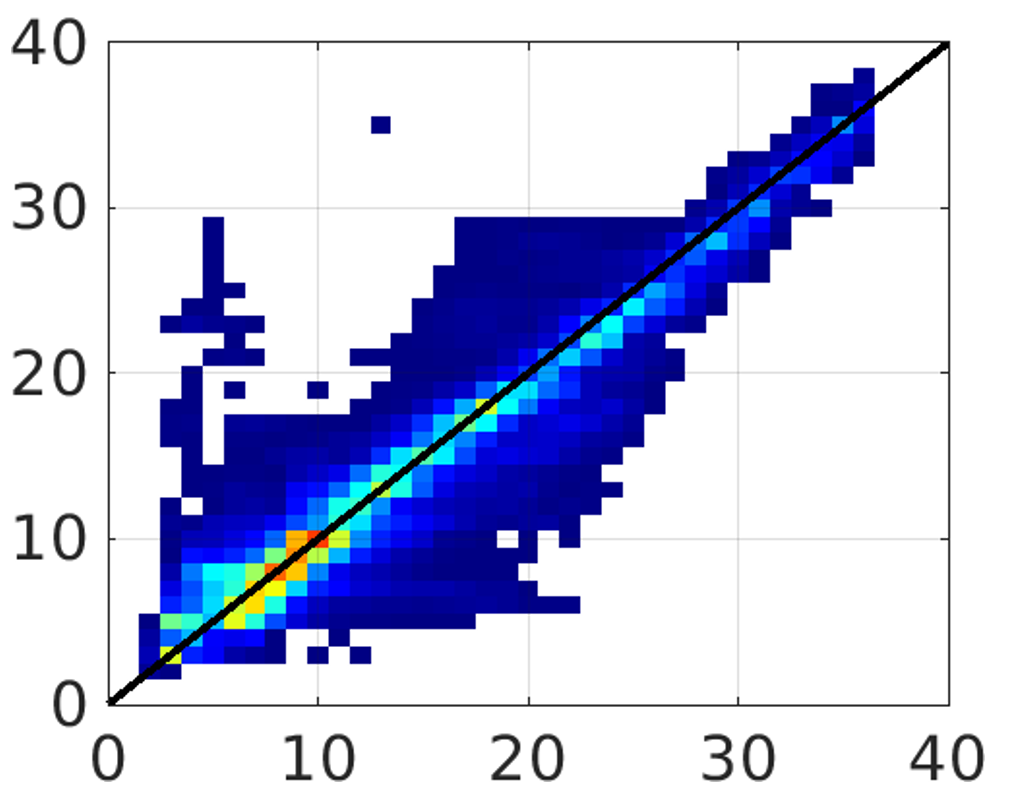} & 
        \includegraphics[width=2.5cm]{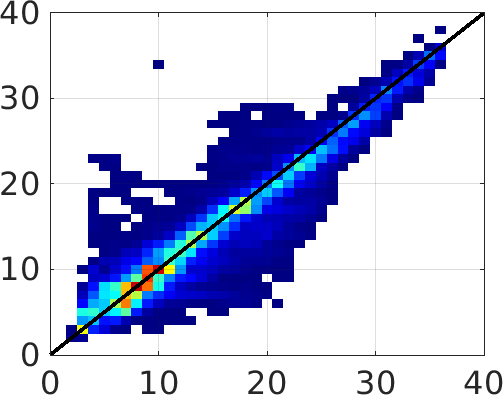} & 
        \includegraphics[width=2.5cm]{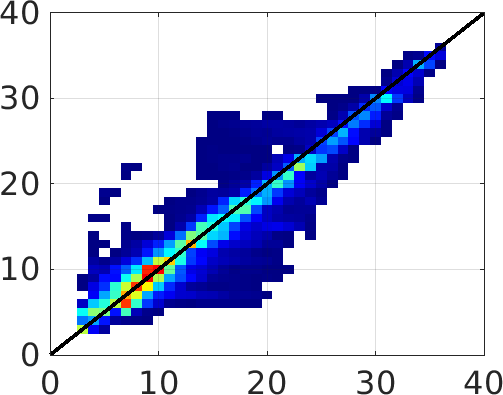} & 
        \includegraphics[width=2.5cm]{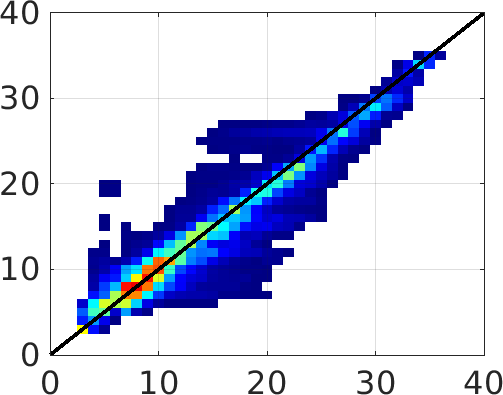} & 
        \includegraphics[width=2.5cm]{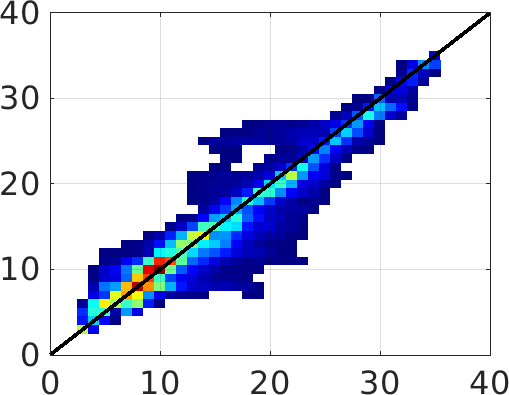} & 
        \includegraphics[width=2.5cm]{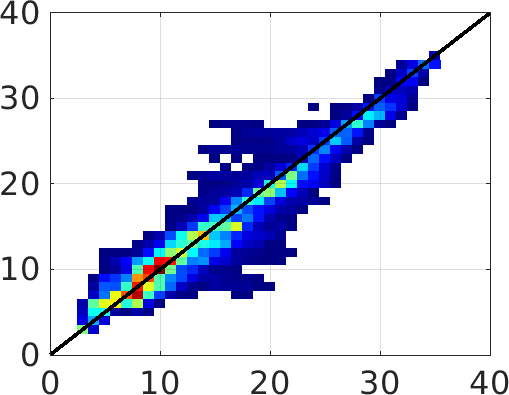} \\

        \multirow{4}{*}[5em]{\makecell{\hspace{-1mm}\rotatebox{90}{\footnotesize Classification-C}}} & 
        \includegraphics[width=2.5cm]{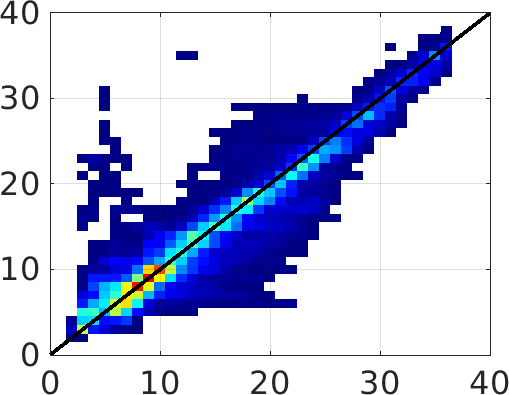} & 
        \includegraphics[width=2.5cm]{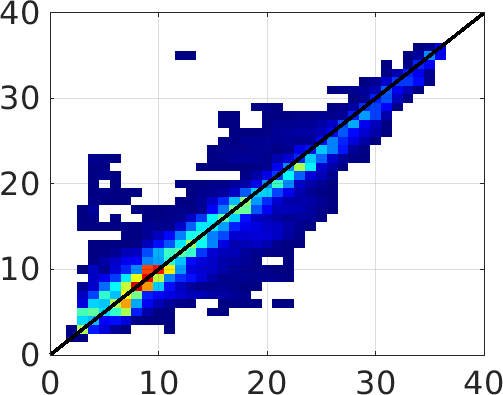} & 
        \includegraphics[width=2.5cm]{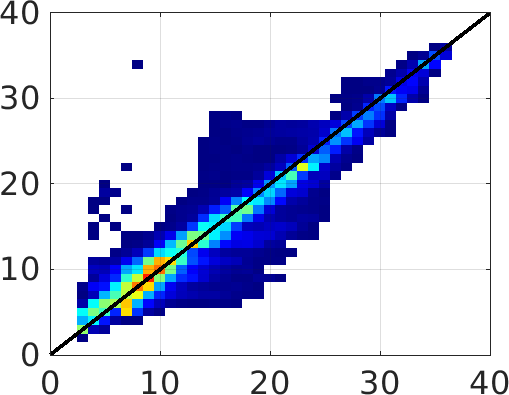} & 
        \includegraphics[width=2.5cm]{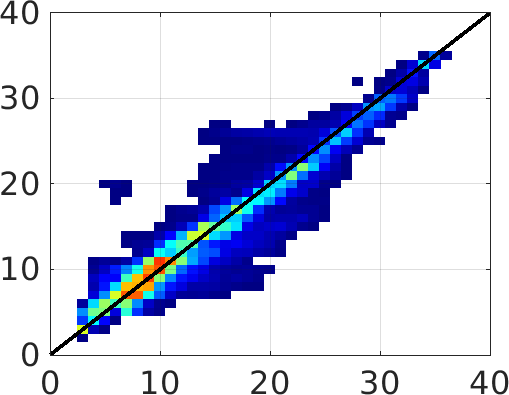} & 
        \includegraphics[width=2.5cm]{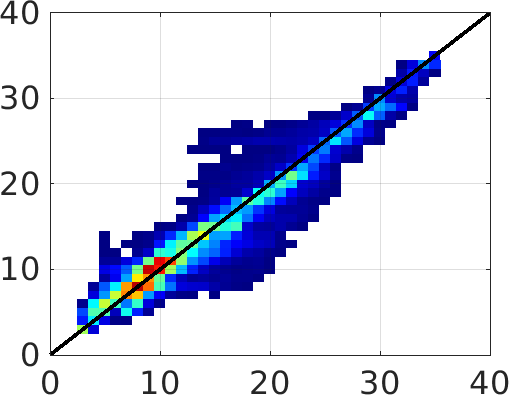} & 
        \includegraphics[width=2.5cm]{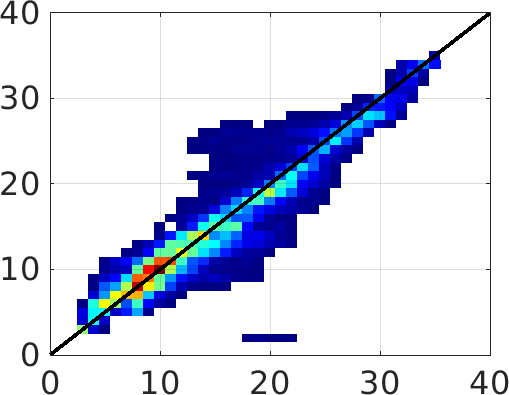} \\

        \multirow{2}{*}[5em]{\makecell{\hspace{-1mm}\rotatebox{90}{\footnotesize Regression-NC}}} & 
        \includegraphics[width=0.15\textwidth]{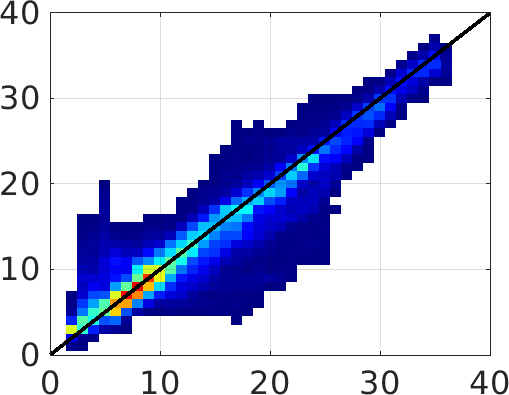} & 
        \includegraphics[width=0.15\textwidth]{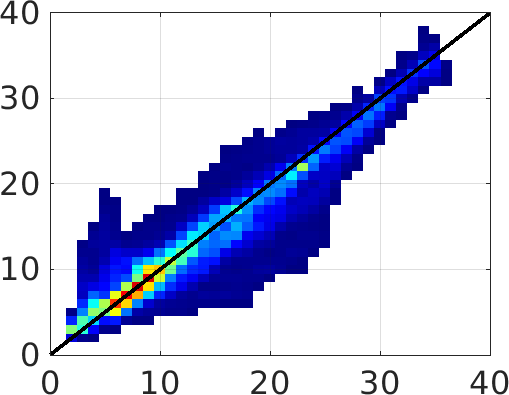} & 
        \includegraphics[width=0.15\textwidth]{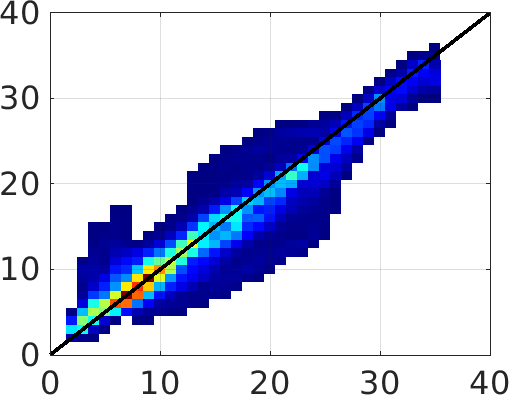} & 
        \includegraphics[width=0.15\textwidth]{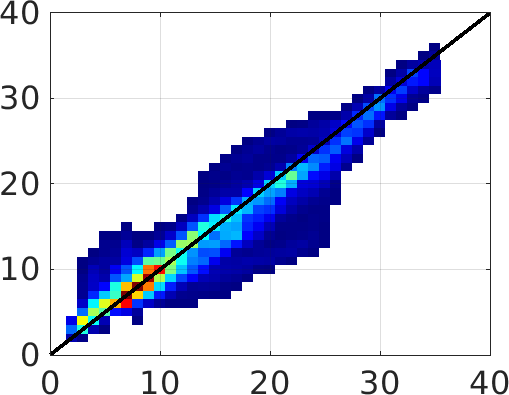} & 
        \includegraphics[width=0.15\textwidth]{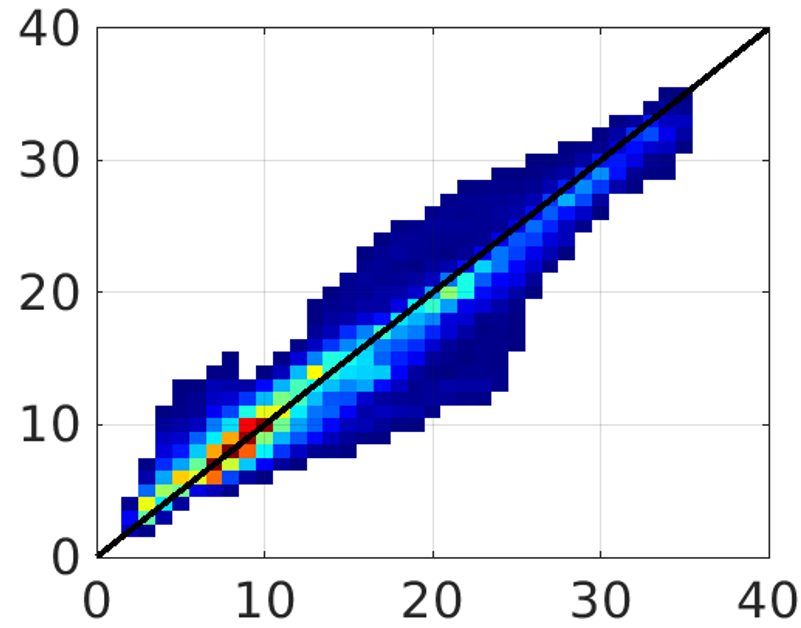} & 
        \includegraphics[width=0.15\textwidth]{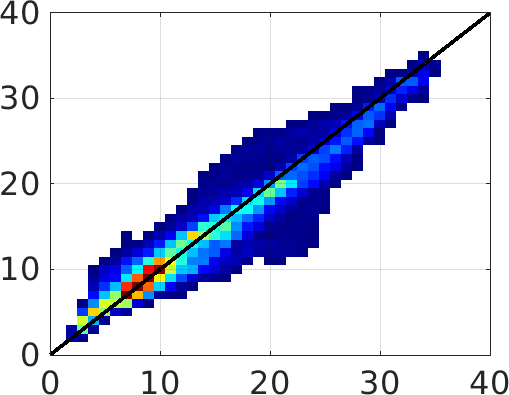} \\
        
        \multirow{2}{*}[5em]{\makecell{\hspace{-1mm}\rotatebox{90}{\footnotesize Regression-C}}} & 
        \includegraphics[width=0.15\textwidth]{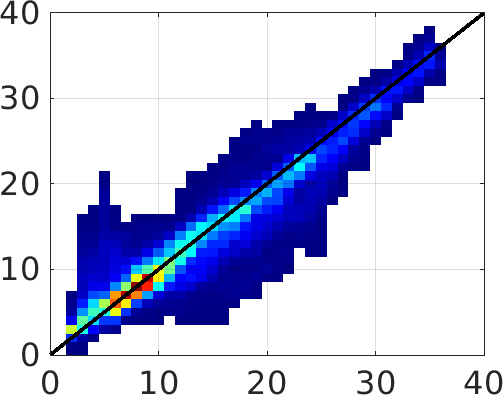} & 
        \includegraphics[width=0.15\textwidth]{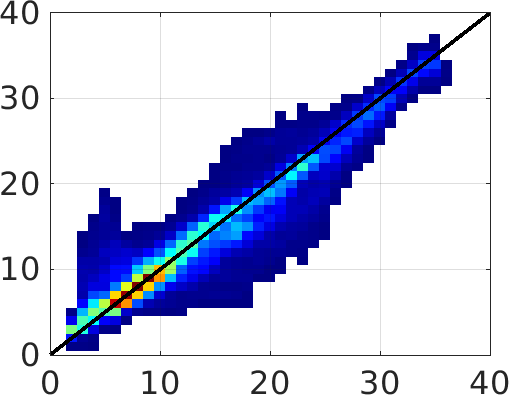} & 
        \includegraphics[width=0.15\textwidth]{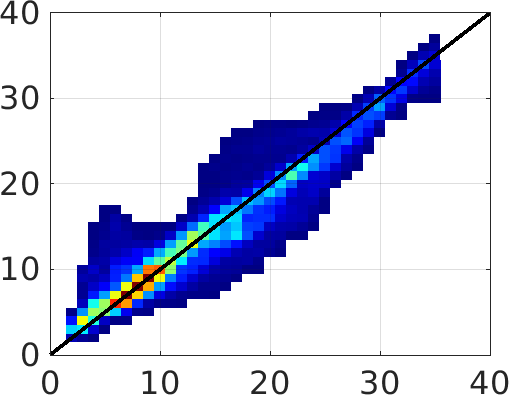} & 
        \includegraphics[width=0.15\textwidth]{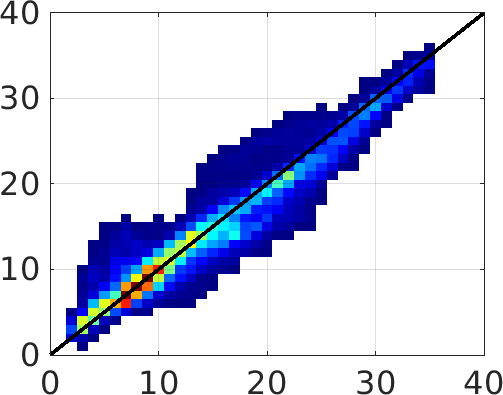} & 
        \includegraphics[width=0.15\textwidth]{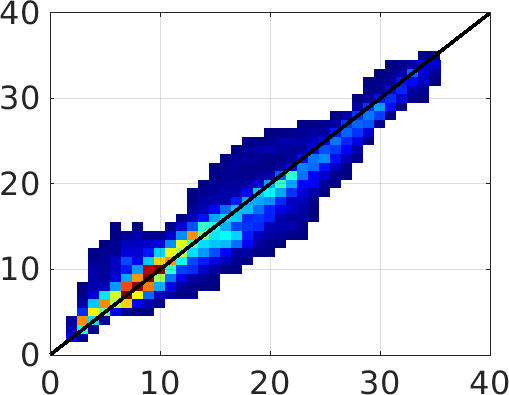} & 
        \includegraphics[width=0.15\textwidth]{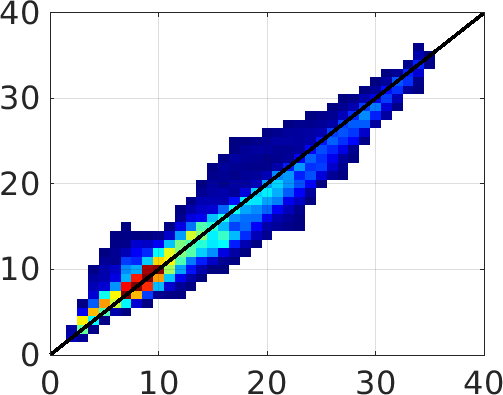} \\
    \end{tabular}
    \caption{Joint distributions between LiDAR-derived DTM and CatBoost predictions. Results are shown for both classification and regression approaches, using C and NC input data, and evaluated across multiple window sizes.
}
    \label{fig: joint_DTM}
\end{figure*}

\begin{figure*}[!h]
    \centering
    
    \begin{tabular}{>{\centering\arraybackslash}m{0cm} c c c c c c}
        &  27$\times$27 &  31$\times$31 & 37$\times$37 &  41$\times$41 &  45$\times$45 &  49$\times$49 \\
         
        
        \multirow{4}{*}[5.4em]{\makecell{\hspace{-1mm}\rotatebox{90}{\footnotesize Classification-NC}}} & 
        \includegraphics[width=2.5cm]{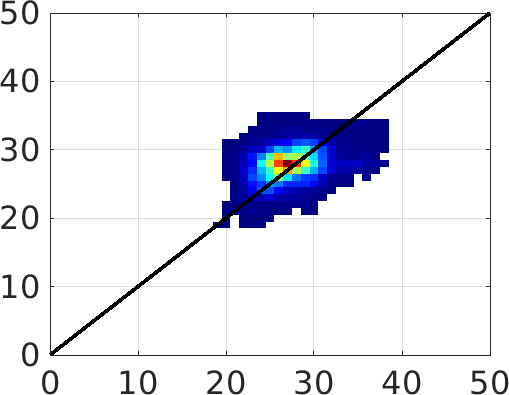} & 
        \includegraphics[width=2.5cm]{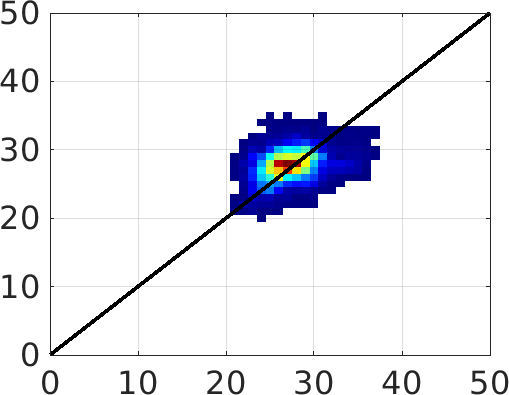} & 
        \includegraphics[width=2.5cm]{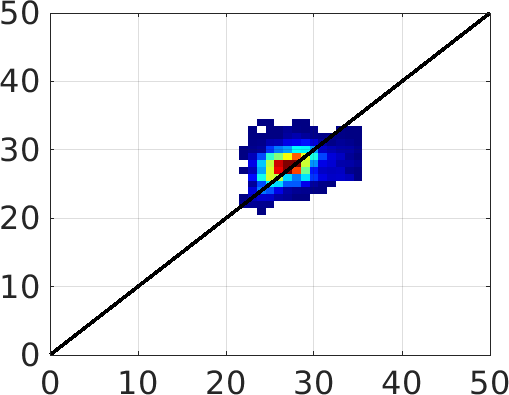} & 
        \includegraphics[width=2.5cm]{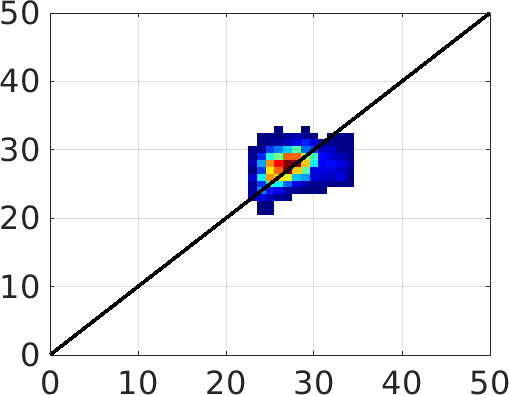} & 
        \includegraphics[width=2.5cm]{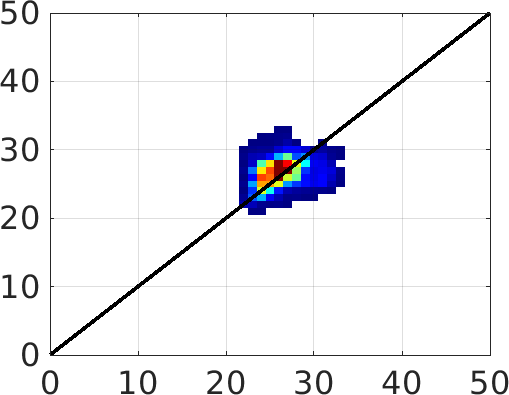} & 
        \includegraphics[width=2.5cm]{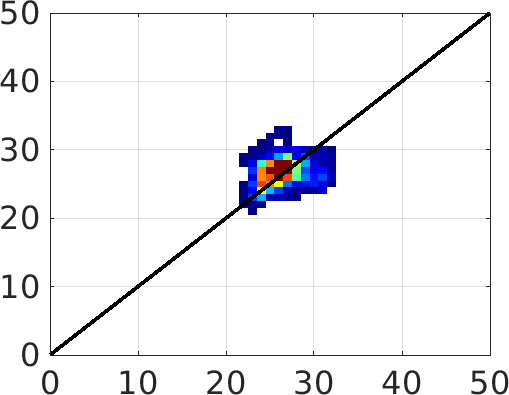} \\

        \multirow{4}{*}[5em]{\makecell{\hspace{-1mm}\rotatebox{90}{\footnotesize Classification-C}}} & 
        \includegraphics[width=2.5cm]{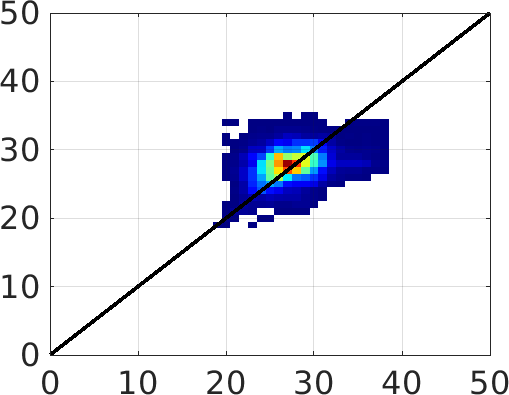} & 
        \includegraphics[width=2.5cm]{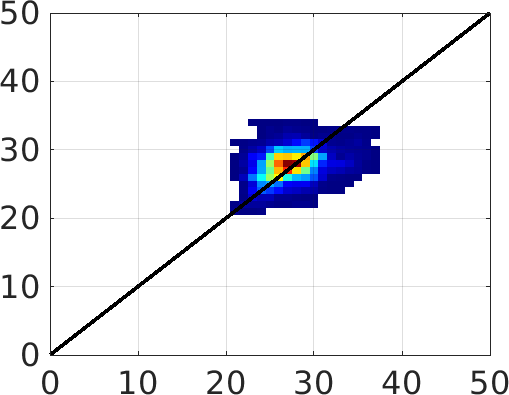} & 
        \includegraphics[width=2.5cm]{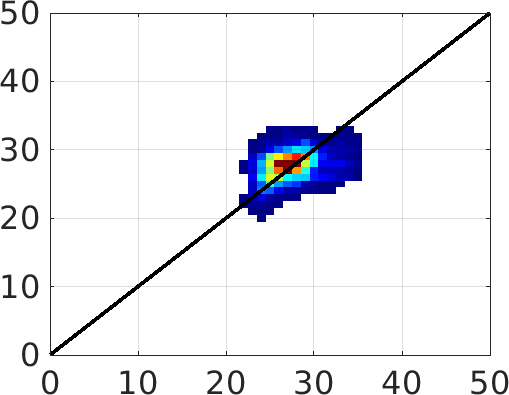} & 
        \includegraphics[width=2.5cm]{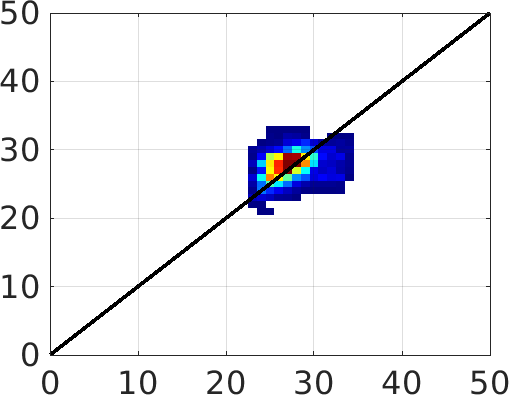} & 
        \includegraphics[width=2.5cm]{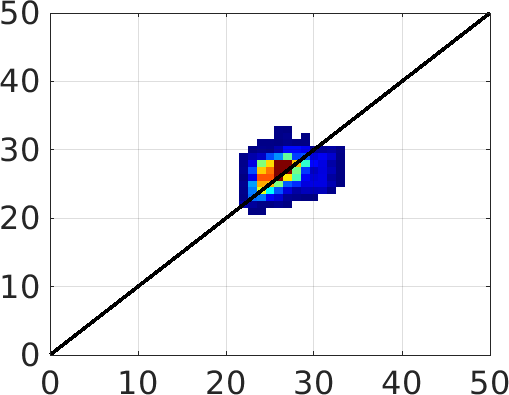} & 
        \includegraphics[width=2.5cm]{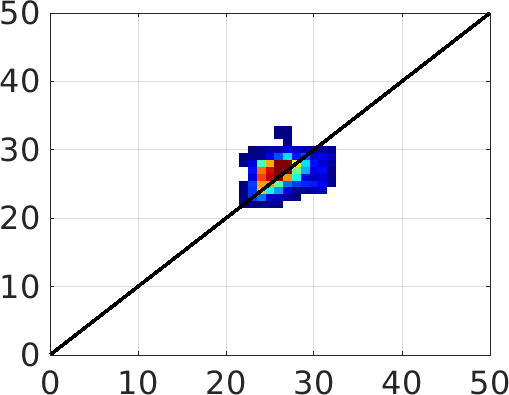} \\

        \multirow{2}{*}[5em]{\makecell{\hspace{-1mm}\rotatebox{90}{\footnotesize Regression-NC}}} & 
        \includegraphics[width=0.15\textwidth]{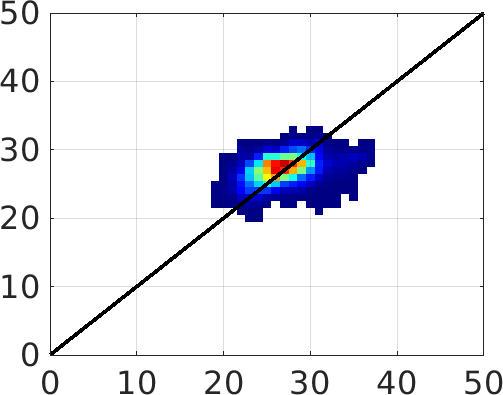} & 
        \includegraphics[width=0.15\textwidth]{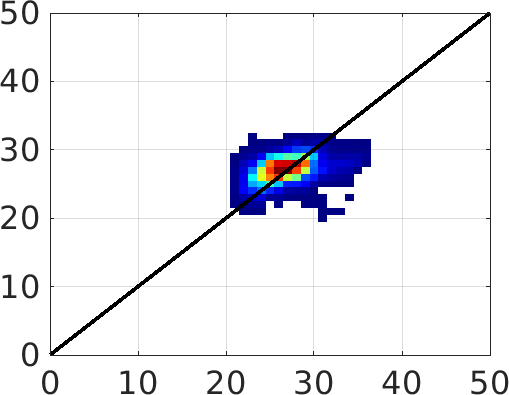} & 
        \includegraphics[width=0.15\textwidth]{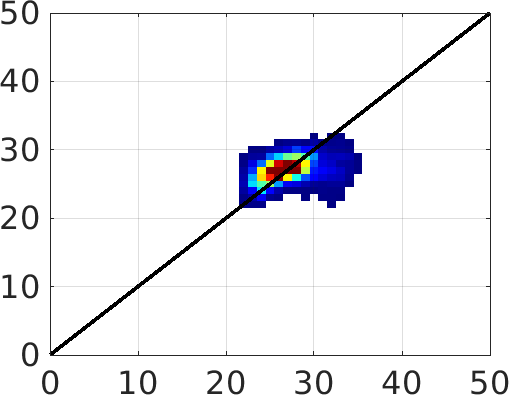} & 
        \includegraphics[width=0.15\textwidth]{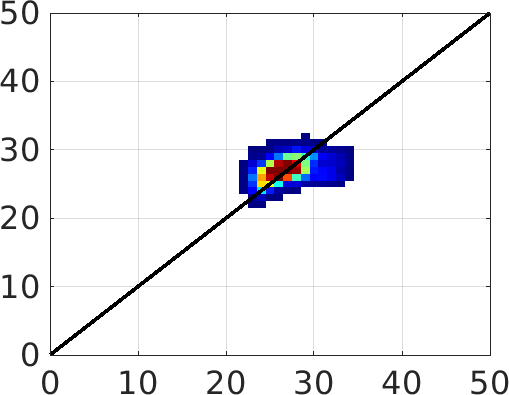} & 
        \includegraphics[width=0.15\textwidth]{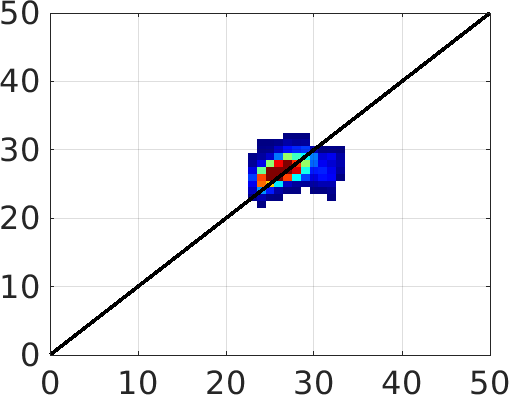} & 
        \includegraphics[width=0.15\textwidth]{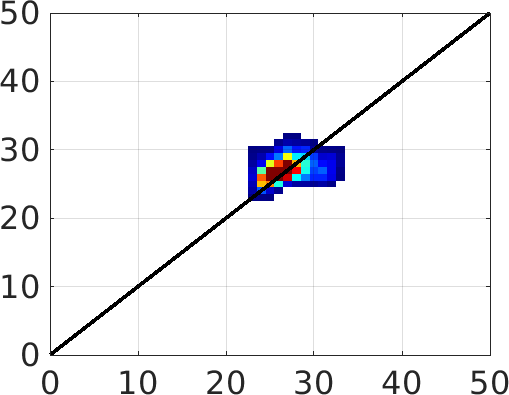} \\
        
        \multirow{2}{*}[5em]{\makecell{\hspace{-1mm}\rotatebox{90}{\footnotesize Regression-C}}} & 
        \includegraphics[width=0.15\textwidth]{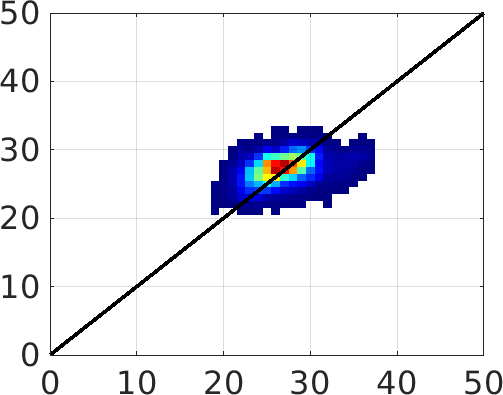} & 
        \includegraphics[width=0.15\textwidth]{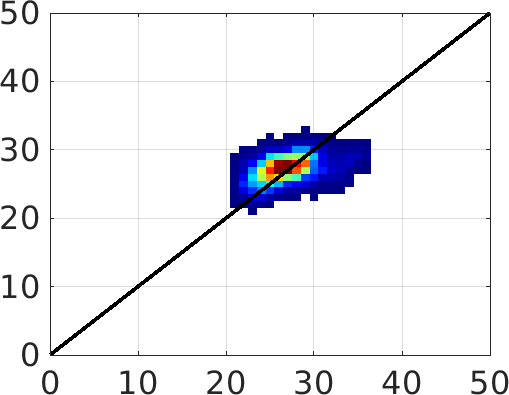} & 
        \includegraphics[width=0.15\textwidth]{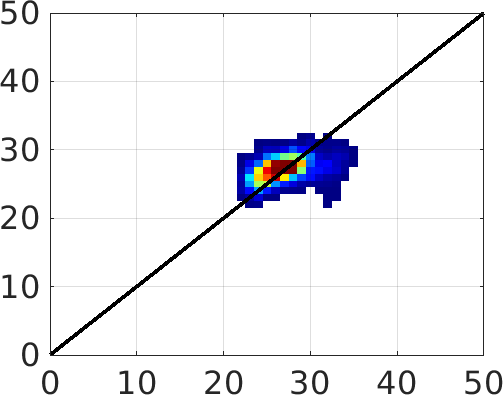} & 
        \includegraphics[width=0.15\textwidth]{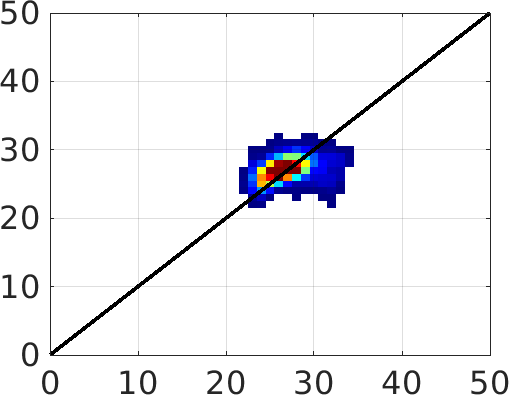} & 
        \includegraphics[width=0.15\textwidth]{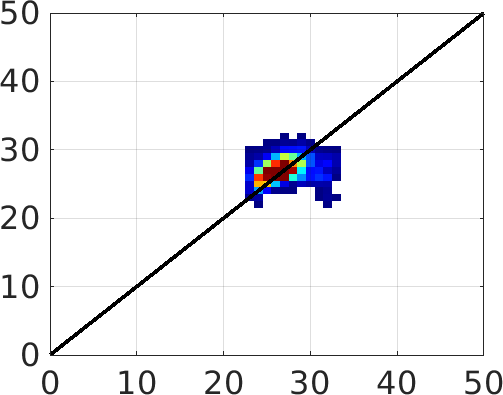} & 
        \includegraphics[width=0.15\textwidth]{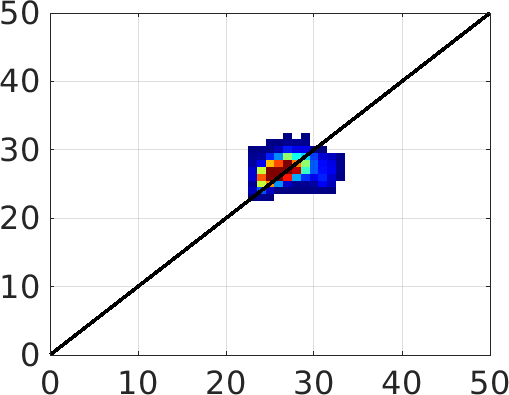} \\
    \end{tabular}
    \caption{Joint distributions between LiDAR-derived CHM and CatBoost predictions. Results are shown for both classification and regression approaches, using C and NC input data, and evaluated across multiple window sizes.}
    \label{fig: joint_CHM}
\end{figure*}

This quantitative evaluation confirms the overall robustness of the  proposed framework across all experimental configurations. As anticipated, increasing the spatial averaging window size generally leads to a reduction in RMSE. However, the improvements are not significant, indicating a relatively low sensitivity of the model to the window size,  as long as the window includes a sufficient number of pixels to estimate the sample covariance matrix.

Regarding data calibration, no significant performance differences are observed between calibrated and non-calibrated inputs. This suggests that the CatBoost-based  algorithm considered in the FGump framework is capable of internally compensating for phase calibration effects, making the explicit calibration step unnecessary. Avoiding this pre-processing operation not only simplifies the pipeline but also reduces the potential for error propagation.

Comparing learning paradigms, the regression-based approach achieves DTM reconstruction quality that is comparable—and in some cases superior—to that of the classification strategy. For CHM estimation, regression consistently yields lower errors. Furthermore, the regression approach eliminates the need for LiDAR data quantization, which is otherwise required for classification, thus preserving more information and reducing preprocessing complexity.
The above observations are confirmed by the qualitative analysis shown in Figures \ref{fig: joint_DTM} and \ref{fig: joint_CHM}, where the joint distribution of preditced DTM and CHM with respect the LiDAR reference are reported. Ideal reconstruction should lie on the bisector line.
Also from the analysis of the Figures \ref{fig: joint_DTM} and \ref{fig: joint_CHM}, it is evident that  the larger the window size, the greater the concentration of predictions on the line. Furthermore, the performance of classification and regression does not differ greatly when applied to calibrated and non-calibrated data.

Based on the above considerations, in the remainder of the paper, we will refer to \textbf{\prop{}} (summarized in the following \textit{Algorithm Box}) as the regression-based implementation using non-calibrated input data and a window size of 49$\times$49, which has shown consistently strong performance across the evaluated metrics.

\RestyleAlgo{ruled}
\LinesNumbered
\begin{algorithm}[!ht]

\captionsetup{labelformat=empty}
\caption{{\textbf{Algorithm Box: FGump}}}

\small

\vspace{2mm}

\tcc{------------- DATA --------------}
MPMB SAR SLC stack ($3N_b$), non-calibrated.

LiDAR-derived CHM or DTM, coregistered to SAR stack.

\vspace{2mm}
\tcc{----------- PROCESSING -----------}
SAR: Compute covariance matrix $\mathbf{R} \in \mathbb{C}^{3N_b \times 3N_b}$ using a spatial window of size $W$ 

LiDAR: Apply spatial averaging to match SAR resolution.

\ForEach{pixel}{
    Extract input feature vector $\mathbf{x} \in \mathbb{R}^{M \times 1}$ from $\mathbf{R}$, where $M = 3N_b + 2(3N_b - 1)$ 
    
    Feed CatBoost Regressor with $\mathbf{x}$ and corresponding LiDAR reference (CHM or DTM)
    }

\vspace{2mm}
\tcc{------------- OUTPUT -------------}
Predicted CHM or DTM values

\end{algorithm}

\subsection{Comparison with other ML approaches} \label{subs: comp_ML}
\begin{figure}[h]
\centering
    \includegraphics[width=8.9cm]{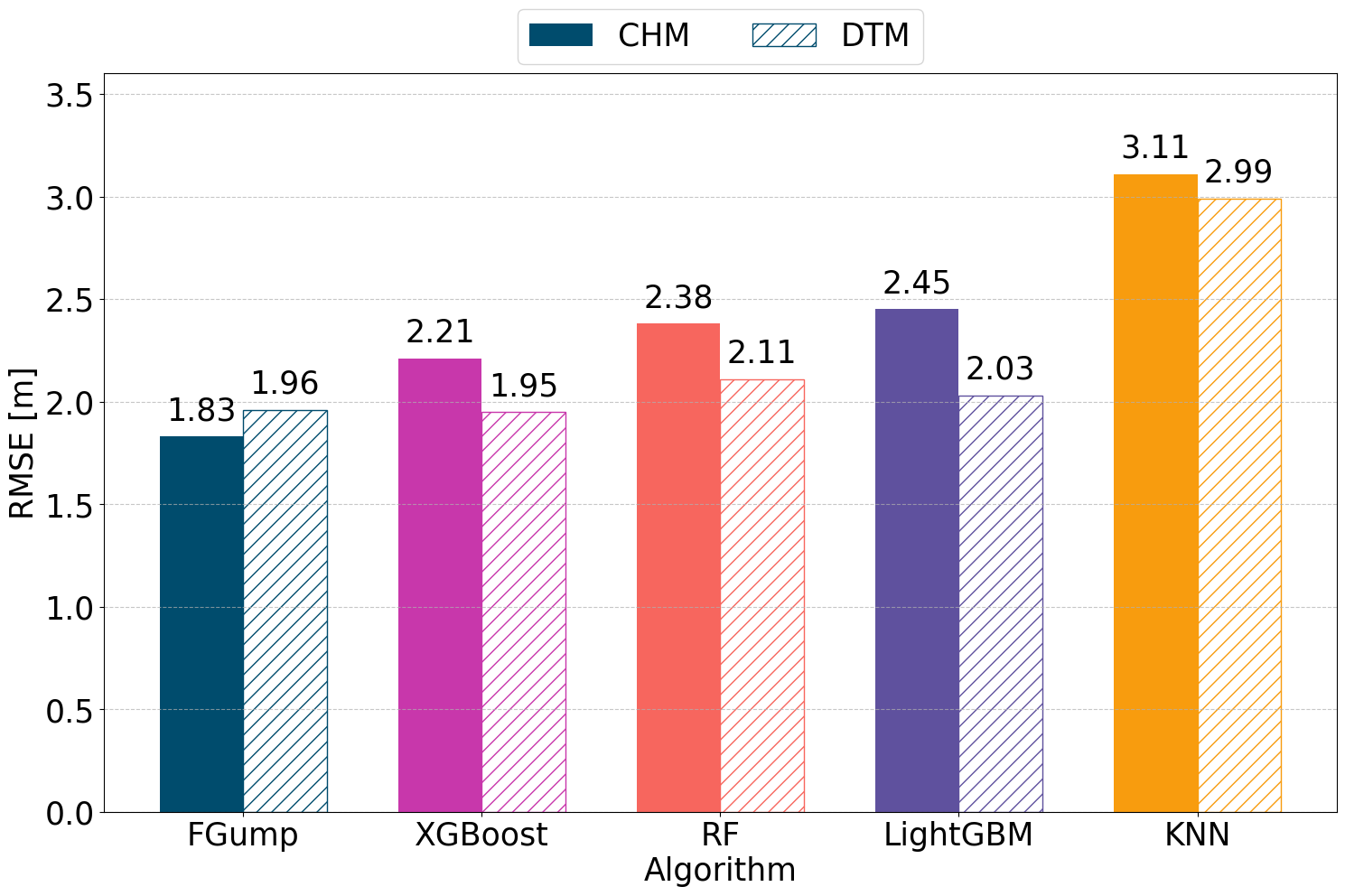}
    \caption{Comparison of RMSE values for DTM and CHM reconstruction using different ML algorithms. CHM results are shown as solid-colored bars, while DTM results are represented with white bars and color-matched diagonal hatching. All models were trained on the same input features and evaluated against LiDAR reference data.
}
    \label{fig: ML comparison}
\end{figure}

To validate the effectiveness of \prop{}, this section presents a comparison with other ML approaches. Following the evaluation framework outlined in \cite{ZHANG2023103532}, the performance of four widely adopted algorithms—Random Forest (RF), XGBoost, LightGBM, and K-Nearest Neighbors (KNN)—is assessed in terms of RMSE, using LiDAR-derived DTM and CHM as reference.
RF is an ensemble method that combines multiple decision trees to enhance predictive accuracy and reduce overfitting. XGBoost and LightGBM, both based on gradient boosting, are designed for high accuracy and computational efficiency, differing primarily in their tree construction strategies and data handling. KNN, on the other hand, is a non-parametric, instance-based method that predicts outcomes based on the similarity to nearby samples. The presented performance evaluation offers a comprehensive perspective on \prop{}’s capabilities in terms of both predictive accuracy and computational efficiency.
The quantitative results are summarized in Fig.~\ref{fig: ML comparison}.

For consistency and fairness, all models were trained using the same input features and reference data as the \prop{} configuration selected in the previous analysis. The results demonstrate the overall superiority of gradient boosting methods. In particular, \prop{} achieves the lowest RMSE for CHM reconstruction, while XGBoost yields the best performance for DTM estimation, with \prop{} providing comparable accuracy.

\subsection{Qualitatively results}
\label{subs: analysis_result}

\begin{figure*}[ht]
    \centering

    \begin{tabular}{ c c c c c c }
         \textbf{LiDAR-CHM} & \textbf{\prop{}} & \textbf{TSNN} & \textbf{XGBoost} & \textbf{SKP} & \textbf{GLRT} \\
        \includegraphics[width=0.15\textwidth]{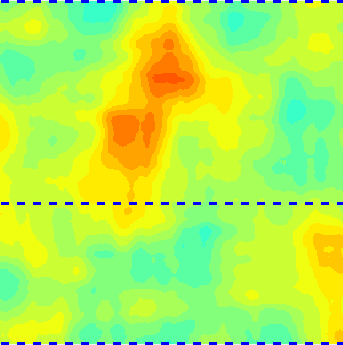} &
        \includegraphics[width=0.15\textwidth]{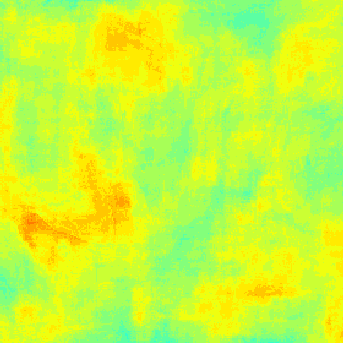} & 
        \includegraphics[width=0.15\textwidth]{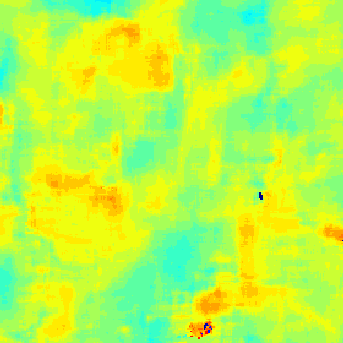} & 
        \includegraphics[width=0.15\textwidth]{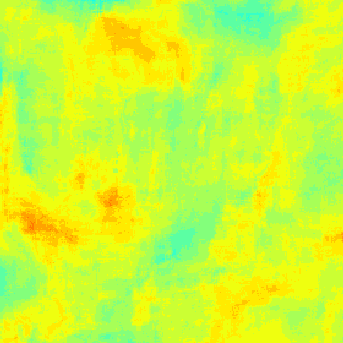} & 
        \includegraphics[width=0.15\textwidth]{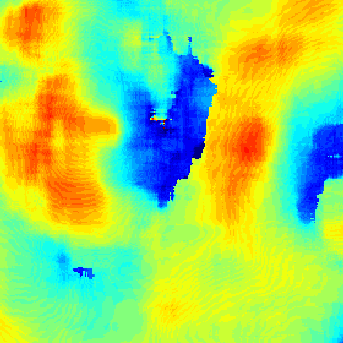} &
        \includegraphics[width=0.15\textwidth]{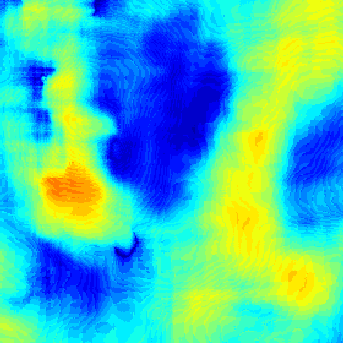} \\
    \end{tabular}

    \vspace{2mm}
    \includegraphics[width=0.3\textwidth]{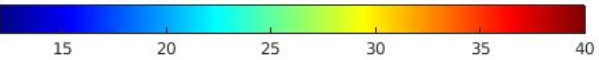}
    \vspace{2mm}

    \begin{tabular}{ c c c c c c }
        \rule{0.15\textwidth}{0pt} & 
        \includegraphics[width=0.15\textwidth]{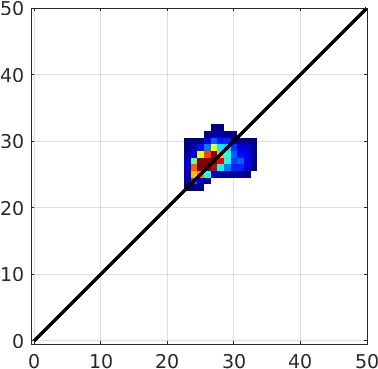} & 
        \includegraphics[width=0.15\textwidth]{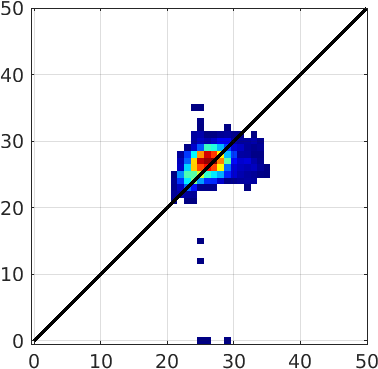} & 
        \includegraphics[width=0.15\textwidth]{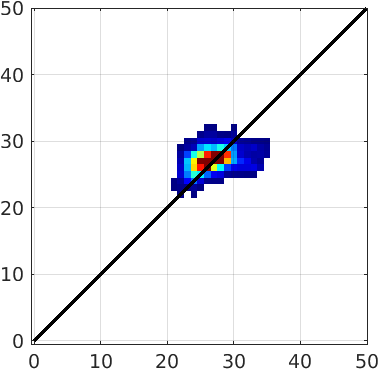} & 
        \includegraphics[width=0.15\textwidth]{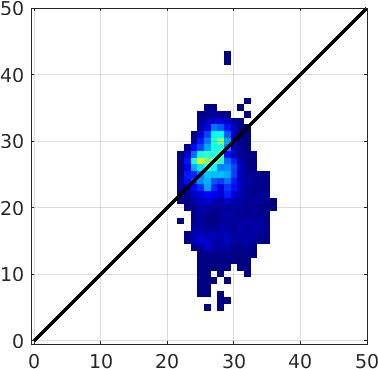} &
        \includegraphics[width=0.15\textwidth]{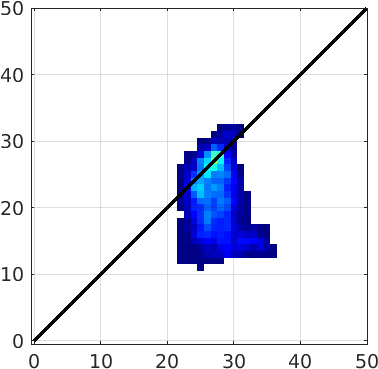} \\
    \end{tabular}

    \caption{Results for CHM reconstruction. From left to right: LiDAR reference, \prop{}, TSNN, XGBoost, SKP, GLRT. CHM height reconstruction in the top row. Colorbar between rows. Joint distribution between reference LiDAR and estimated heights in the last row.}
    \label{fig:CHM_results}
\end{figure*}

\begin{figure*}[!ht]
    \centering
    \begin{tabular}{c@{\hskip 0.1cm}c@{\hskip 0.1cm}c@{\hskip 0.1cm}c}
        \rotatebox{90}{\makebox[5.1cm]{\textbf{Height [m]}}} &
        \includegraphics[width=0.31\textwidth]{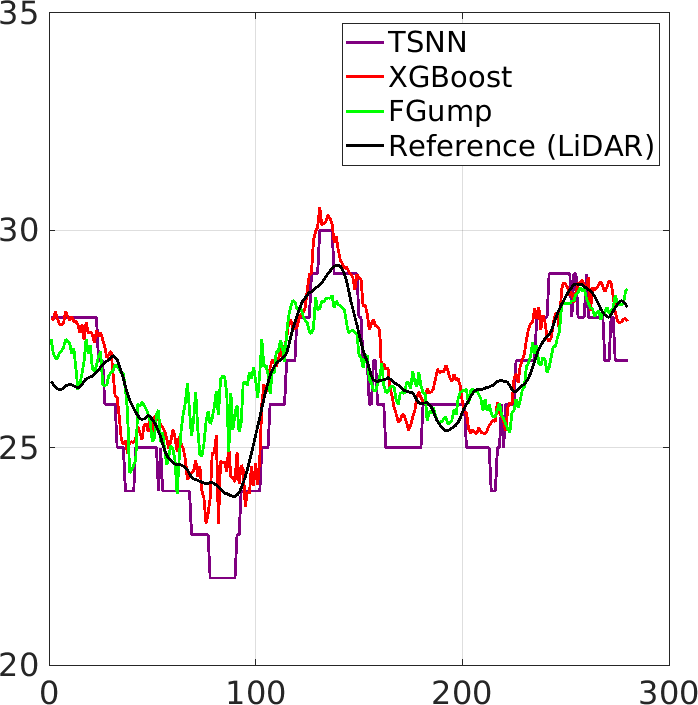} & 
        \includegraphics[width=0.31\textwidth]{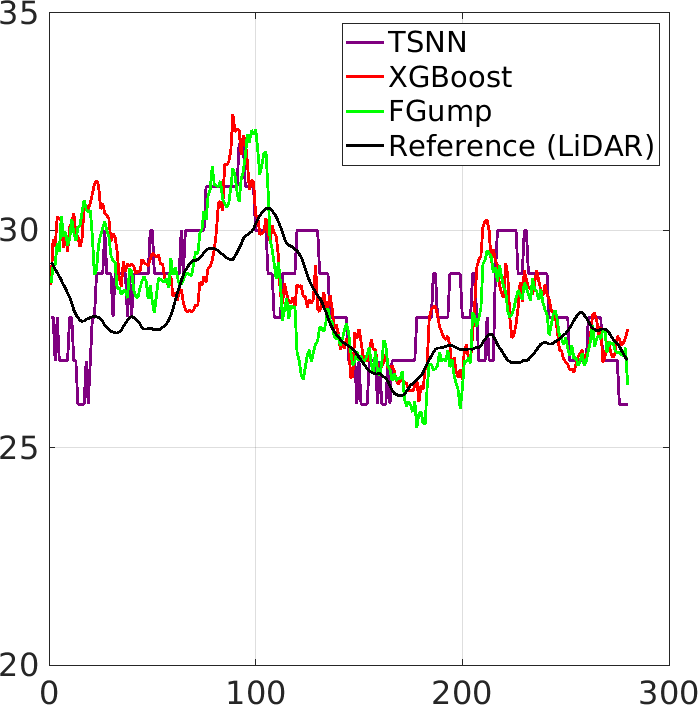} &
        \includegraphics[width=0.31\textwidth]{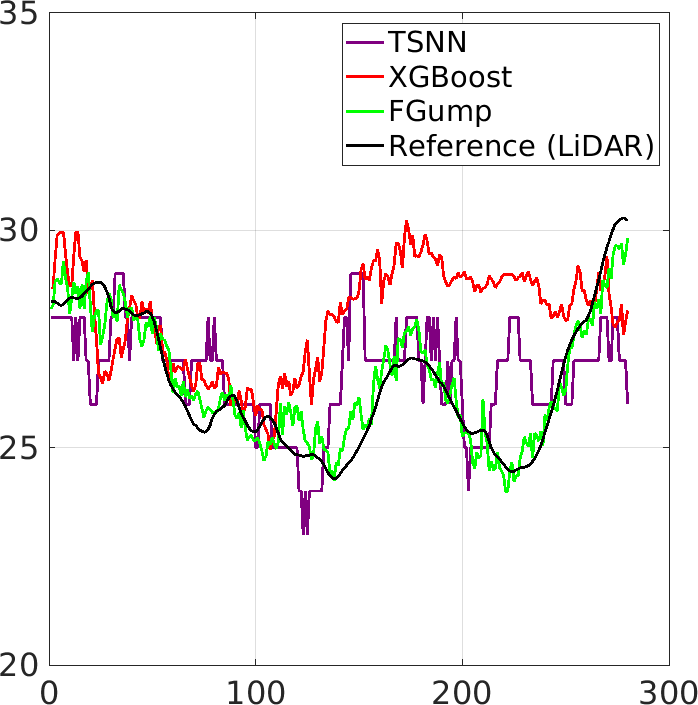} \\
        & \makebox[0.25\textwidth]{\textbf{Pixel}} & 
        \makebox[0.25\textwidth]{\textbf{Pixel}} &
        \makebox[0.25\textwidth]{\textbf{Pixel}} \\
    \end{tabular}
    \caption{Tracelines for CHM reconstruction: LiDAR reference (black), \prop{} in green, XGBoost in red, TSNN in purple}
    \label{fig:traceline_chm}
\end{figure*}

\begin{figure*}[!ht]
    \centering

    \begin{tabular}{ c c c c c c }
         \textbf{LiDAR-DTM} & \textbf{\prop{}} & \textbf{TSNN} & \textbf{XGBoost} & \textbf{SKP} & \textbf{GLRT} \\
        \includegraphics[width=0.15\textwidth]{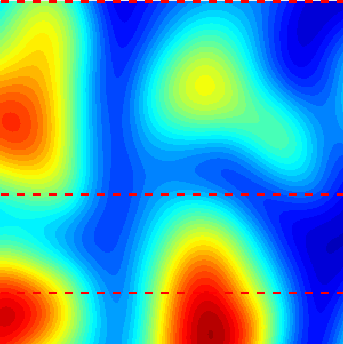} &
        \includegraphics[width=0.15\textwidth]{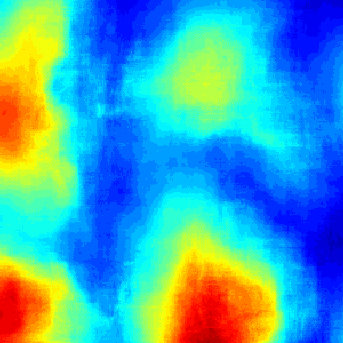} & 
        \includegraphics[width=0.15\textwidth]{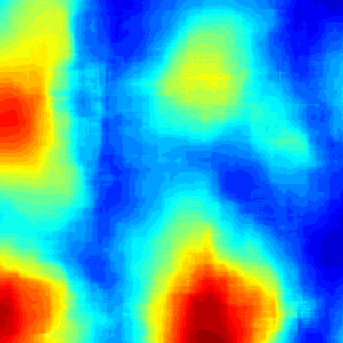} & 
        \includegraphics[width=0.15\textwidth]{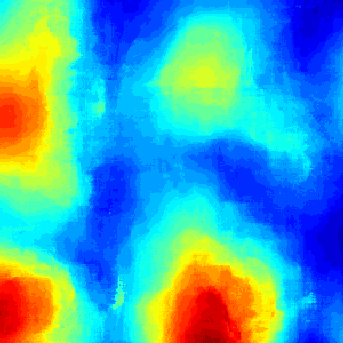} & 
        \includegraphics[width=0.15\textwidth]{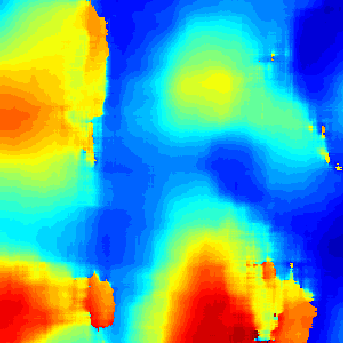} &
        \includegraphics[width=0.15\textwidth]{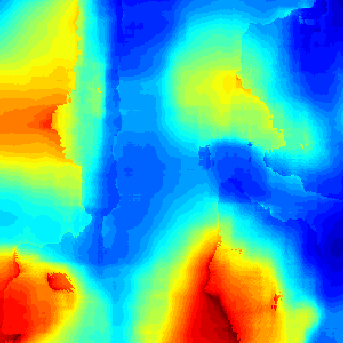} \\
    \end{tabular}

    \vspace{2mm}
    \includegraphics[width=0.3\textwidth]{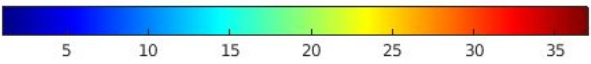}
    \vspace{2mm}

    \begin{tabular}{ c c c c c c }
        \rule{0.15\textwidth}{0pt} &  
        \includegraphics[width=0.15\textwidth]{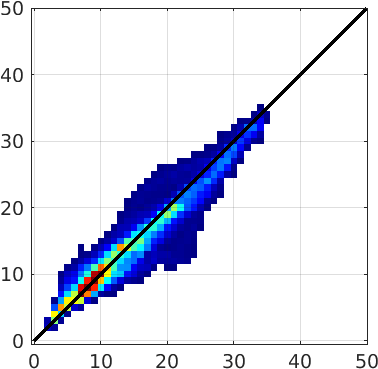} & 
        \includegraphics[width=0.15\textwidth]{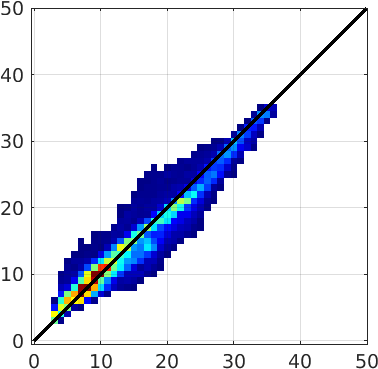} & 
        \includegraphics[width=0.15\textwidth]{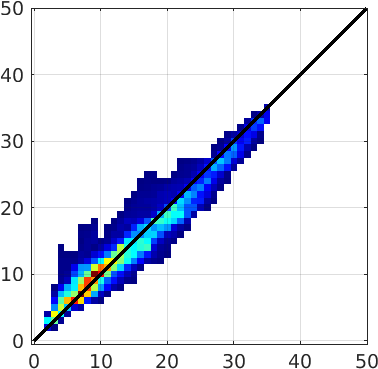} & 
        \includegraphics[width=0.15\textwidth]{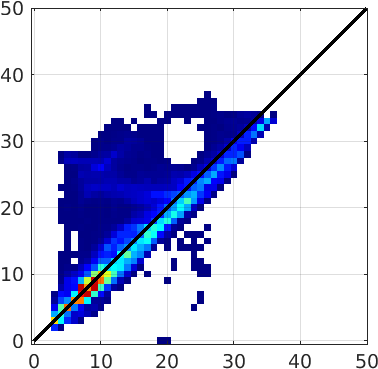} &
        \includegraphics[width=0.15\textwidth]{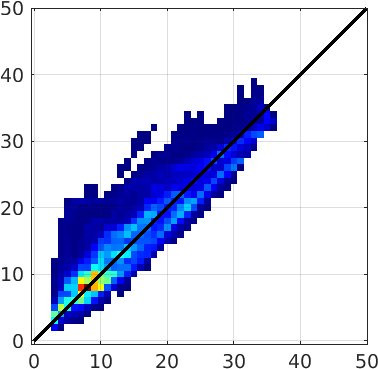} \\
    \end{tabular}

    \caption{Results for DTM reconstruction. From left to right: LiDAR reference, \prop{}, TSNN, XGBoost, SKP, and GLRT. DTM height reconstruction in the top row. Joint distribution between reference LiDAR and estimated heights in the last row.}
    \label{fig:DTM_results}
\end{figure*}

\begin{figure*}[!h]
    \centering
    \begin{tabular}{c@{\hskip 0.1cm}c@{\hskip 0.1cm}c@{\hskip 0.1cm}c}
        \rotatebox{90}{\makebox[5.1cm]{\textbf{Height [m]}}} &
        \includegraphics[width=0.298\textwidth]{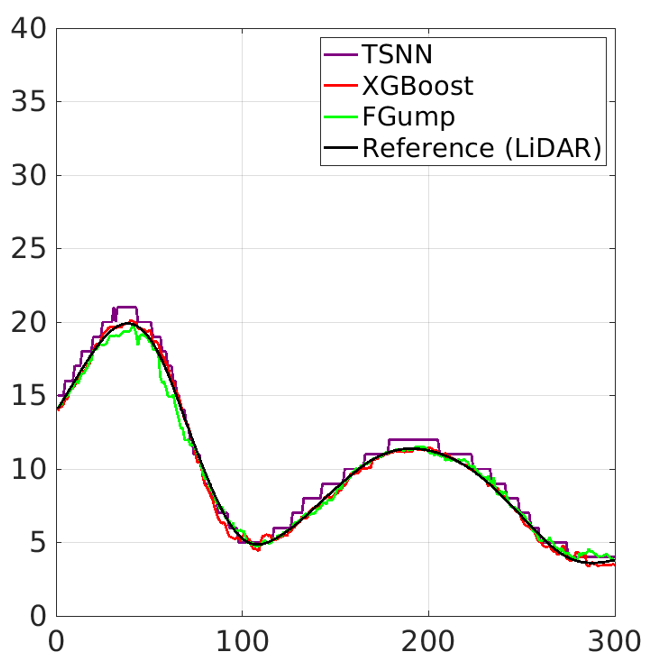} & 
        \includegraphics[width=0.291\textwidth]{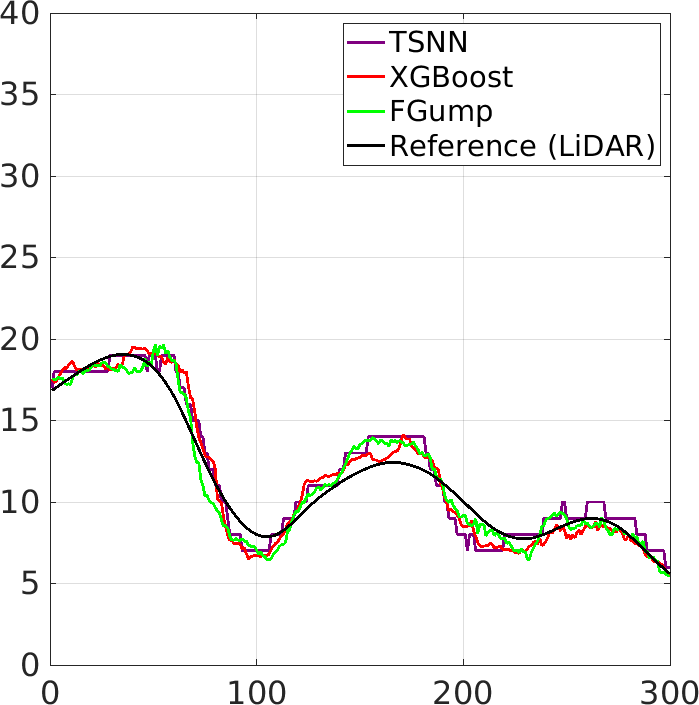} &
        \includegraphics[width=0.291\textwidth]{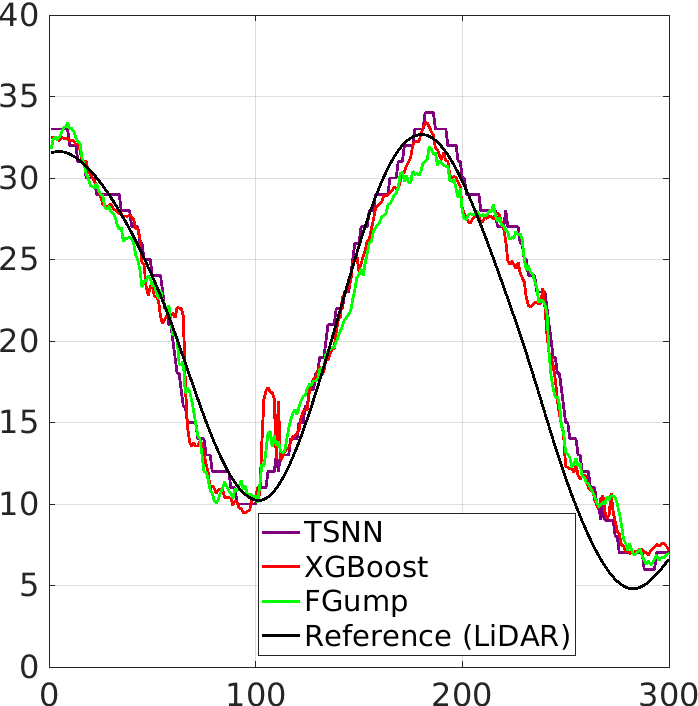} \\
        & \makebox[0.25\textwidth]{\textbf{Pixel}} & 
        \makebox[0.25\textwidth]{\textbf{Pixel}} &
        \makebox[0.25\textwidth]{\textbf{Pixel}} \\
    \end{tabular}
    \caption{Tracelines for DTM reconstruction: LiDAR reference (black), \prop{} in green, XGBoost in red, TSNN in purple}
    \label{fig:traceline_dtm}
\end{figure*}

In this Section the comparison with SOTA solutions is carried out: two data-driven solutions such as TSNN and XGBoost, and two traditional methods such as the Sum of Kronecker Products (SKP) decomposition \cite{ImSKP29} with Capon beamforming and the Generalized Likelihood Ratio Test (GLRT) \cite{Hossein}, have been considered. TSNN and XGBoost have been retrained using the same training dataset as the proposal. SKP decomposition is a classical technique that decomposes the covariance matrix into the sum of several Kronecker products and retains the first two terms as the ground and volume contributions. A spectral reconstruction method, such as Capon beamforming, is applied to these contributions, and the peaks of the reconstructed profiles indicate the heights.  The GLRT method is based on a statistical detection approach using the generalized likelihood ratio test to detect ground and canopy scattering along the vertical direction. Consequently, the height is determined at the phase center location where most of the backscattered power is concentrated. However, in the P-band case, the height corresponds to the center of the distribution loss zone, as demonstrated in \cite{Tebaldini2012}. Therefore, in our experiments, an offset is applied to the tomographic reconstructions using the LiDAR-based data to compensate for the height mislocation.
A qualitative and quantitative assessment is reported for a fully evaluation of the results. Due to the offset issue, the comparison with SKP and GLRT is limited to a qualitative assessment.

In Fig. \ref{fig:CHM_results} the visual reconstruction of the CHM testing patch provided by comparison method, alongside the LiDAR reference one, and the resulting joint distributions are reported. Generally, as shown by joint distributions, the proposed solution, together with TSNN and XGBoost, show a good agreement with LiDAR reference, while SKP
and GLRT are suffering of an quite evident under-estimation effect. 

To better highlight the differences in the CHM profile reconstructions of TSNN, XGBoost and \prop{}, which from Fig. \ref{fig:CHM_results} seem to achieve a comparable performance, Fig. \ref{fig:traceline_chm} presents three CHM reconstruction tracelines extracted along three horizontal lines of the image at row indices 1, 160, and 280. The LiDAR-derived CHM (black) serves as the reference, while the predicted profiles of \prop{} (green), XGBoost (red), and TSNN (purple) are overlaid. Visual inspection reveals that \prop{} consistently provides a smoother and more stable approximation of the reference signal, closely following the LiDAR profile even in regions of high spatial variability. XGBoost captures the general shape of the canopy structure but tends to introduce overestimations in peak regions. TSNN, on the other hand, exhibits frequent oscillations and discontinuities, particularly in flat or dense vegetation areas, indicating limited generalization. These qualitative trends align with the quantitative results, confirming the robustness and adaptability of \prop{} in reconstructing fine-grained CHM features. Notably, the traceline validation is computed over more than one million points, ensuring statistical significance.


Similar consideration can be drawn for the DTM prediction. In Fig. \ref{fig:DTM_results}, the prediction of DTM provided by \prop{}, TSNN, XGBoost, SKP and GLRT, together with the LiDAR reference and corresponding joint-distribution are shown. According to the joint-distributions, the data-driven approaches provide predictions distributed along the bisector, with TSNN slightly more focused. 
Then, it can be stated that TSNN, XGBoost and \prop{} achieve similar performance in DTM estimation, while the traditional methods, SKP and GLRT, show a marked over-estimation.

As made for CHM, to better highlight the differences in the DTM profile reconstructions of TSNN, XGBoost and \prop{}, Fig. \ref{fig:traceline_dtm} shows the qualitative performance of the DTM reconstruction through tracelines extracted along image rows 1, 170, and 256. The LiDAR-derived DTM (black line) serves as the reference, while the outputs of \prop{} (green), XGBoost (red), and TSNN (purple) are superimposed. All three models approximate the terrain surface with a high degree of consistency. However, \prop{} most closely follows the LiDAR profile, especially in gently varying topographies. XGBoost performs comparably, though it occasionally overshoots at elevation peaks. TSNN, while able to reproduce general trends, exhibits local mismatches.

\begin{figure}
    \centering
    \includegraphics[width=8.7cm]{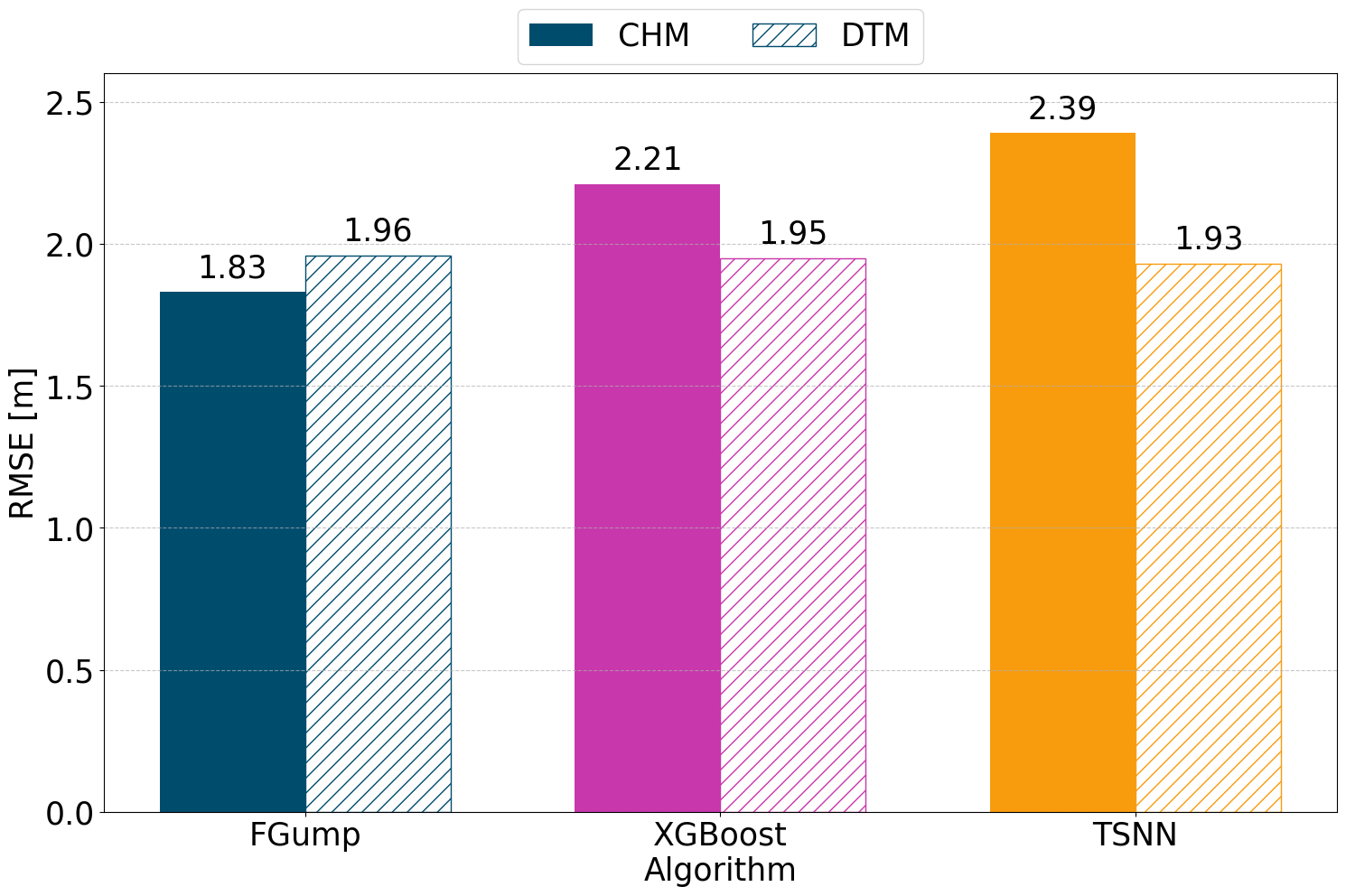}
    \caption{{Comparison of RMSE} values for CHM and DTM reconstruction using \prop{}, XGBoost, and TSNN. CHM results are shown as solid-colored bars, while DTM results are depicted with white bars and color-matched hatching. All models were trained on the same input features and evaluated against LiDAR reference data.
}
    \label{tab: numerical comparison}
\end{figure}

The numerical analysis reported in Fig. \ref{tab: numerical comparison} confirms the previous consideration: \prop{} outperforms all the methods in the reconstruction of the CHM profile, while all the data-driven approaches are almost similar each-other for the reconstruction of the DTM profile.

\subsection{Computational Time} \label{subs: compcost}
Another important result of the proposed framework is the impact on the computational time.  In scientific research and practical applications, achieving low training times for ML models is of paramount importance. Faster training times enable more efficient model development and iteration, allowing researchers and practitioners to test and refine ideas quickly, thereby accelerating progress. More importantly, in case of the forest monitoring, having a low training time is of fundamental importance: applying the same solution to different areas, different data, and different forest, may benefit of a fine-tuning step \cite{Yang2024Igarss}.

\begin{figure}
    \includegraphics[width=1.05 \columnwidth]{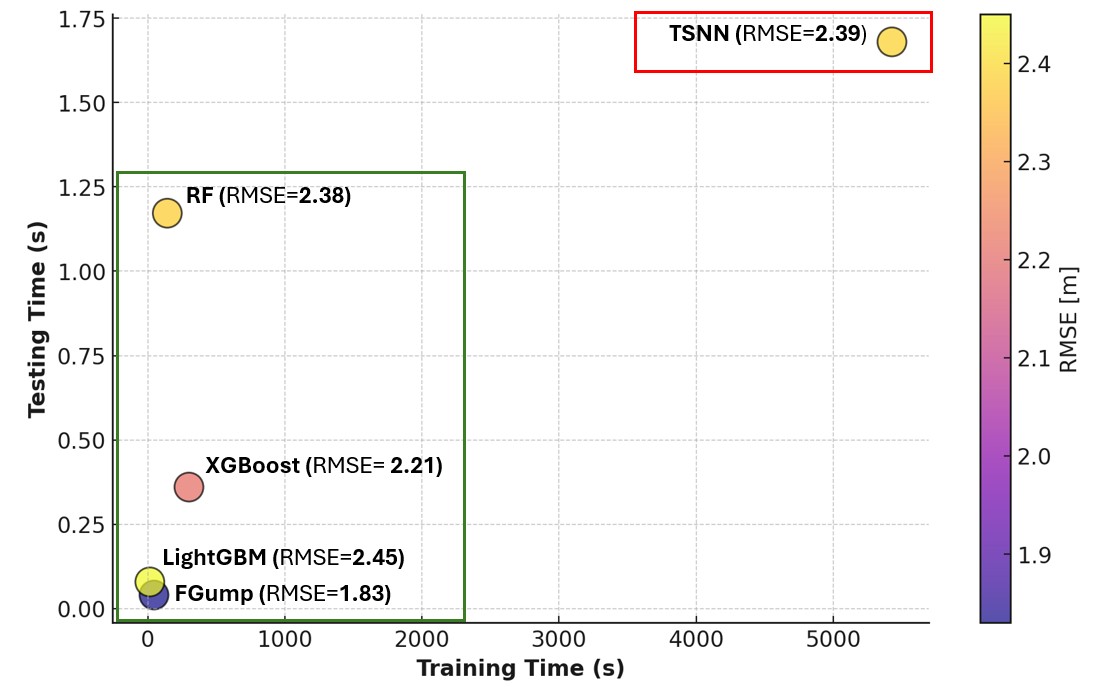}
    \caption{Computational performance comparison among \prop{}, XGBoost, LightGBM, RF and TSNN. }
    \label{fig: computational time}
\end{figure}

Indeed, in Fig. \ref{fig: computational time}, a comparison of the computational time for the training and testing phases of the data-driven approaches is reported excepted for KNN whose testing time is higher then other methods (9.56s). The experiments have been carried-out on on a GeForce GTX 1080Ti GPU with 12 GB of memory. 
For this analysis the training and testing dataset for CHM reconstruction is considered. 
As shown in the figure, \prop{} has the most favorable profile, achieving the shortest training time (46~s) and lowest testing time (0.04~s), with a moderate model size (3~$\times$~10\textsuperscript{6} parameters). XGBoost, while requiring a shorter training time than TSNN (301~s vs. 5428~s), has the highest number of parameters (4.7~$\times$~10\textsuperscript{6}) and a longer inference time (0.36~s). TSNN, despite having fewer parameters than XGBoost (2.5~$\times$~10\textsuperscript{6}), results in the highest computational burden, with substantially longer training and testing times (5428~s and 1.68~s, respectively). These results confirm the computational advantages of \prop{}, making it a practical and scalable solution for operational applications.

\section{Conclusion} \label{sec: conclusion}
 
This study presents the FGump framework as an efficient and accurate solution for forest height estimation using SAR data. By utilizing a limited set of hand-designed features and avoiding complex pre-processing (e.g., calibration and/or quantization), FGump strikes an effective balance between accuracy and computational efficiency. The regression-based approach enables continuous, fine-grained estimations, eliminating quantization artifacts and providing precise measurements.
Experimental results show that FGump, trained on non-calibrated data with an optimized spatial window, consistently outperforms SOTA ML and DL methods in both CHM and DTM reconstruction tasks. Furthermore, FGump achieves high predictive accuracy with significantly reduced computational costs compared to more complex architectures, such as TSNN, and traditional TomoSAR algorithms like SKP and GLRT.
The framework’s ability to operate without heavy pre-processing while maintaining strong generalization makes it particularly suitable for large-scale, time-sensitive forestry applications, offering a practical alternative for operational integration of ML-based height mapping systems using SAR data.

\section*{Acknowledgment}
This work was (partially) supported by the European Union under the Italian National Recovery and Resilience Plan (PNRR) of NextGenerationEU 
within the Project MANSAR-HPC - Monitoraggio Ambientale di aree Naturali tramite tecnologia SAR e sistemi HPC - call “Bando a Cascata”, D. R. Università degli Studi di Bari Aldo Moro n. 1433 del 17/04/2024, Project “HPC – National Centre for HPC, Big Data and Quantum Computing”, Code CN00000013, CUP H93C22000450007, Spoke 5 “Environment \& Natural Disasters”, call MUR n. 3138 16/12/2021 PNRR, Missione 4, Componente 2, Investimento 1.4– NextGenerationEU.

\newpage
    
    \bibliographystyle{IEEEtran}
    \bibliography{IEEEabrv}

    \begin{IEEEbiography}[{\includegraphics[width=1in,height=1.15in,clip,keepaspectratio]{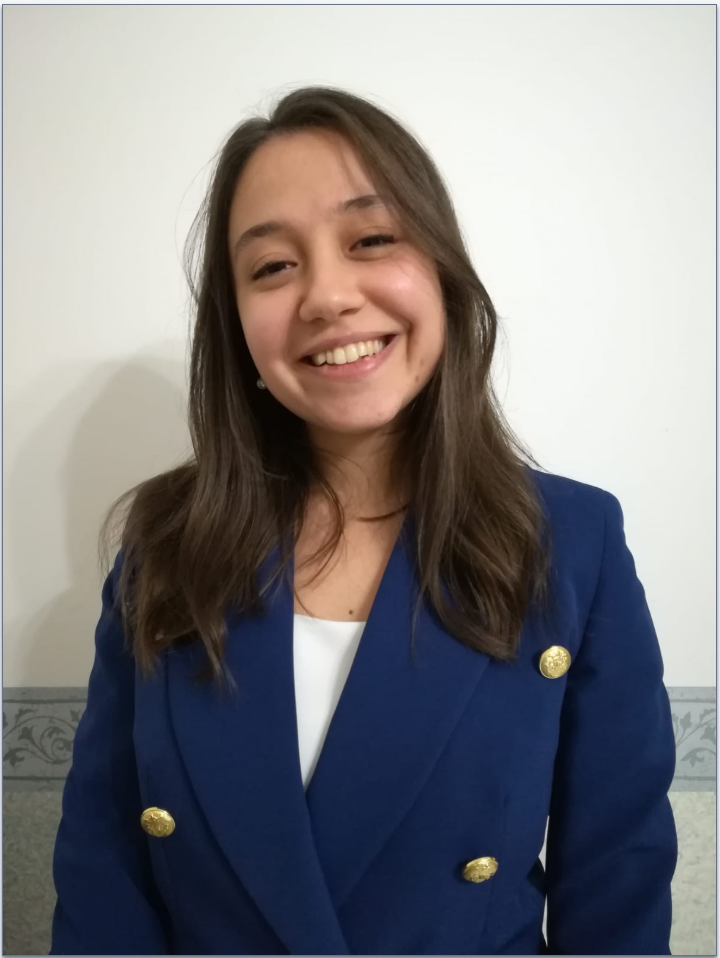}}]{Francesca Razzano}
    graduated cum laude in Electronic Engineering for Automation and Telecommunications from the University of Sannio in 2023. She is currently enrolled in the Ph.D. program in Information and Communication Technology and Engineering at the University of Parthenope, in Naples, under the supervision of Prof. Gilda Schirinzi and co-supervision of Prof. Silvia L. Ullo. Her research focuses primarily on Remote Sensing and satellite data analysis, as well as the application of Artificial Intelligence techniques for Earth observation. In particular, she investigates water quality monitoring and forest tree height estimation. She has also contributed to research on the fusion of optical and SAR data for different tasks. In addition, she works on the development of onboard AI for Remote Sensing systems. Her professional experience includes a position as a Visiting Researcher at the European Space Agency’s $\Phi$-Lab. She has co-authored papers and articles presented at renowned conferences in the field of remote sensing. She is an IEEE Student Member, actively involved in IEEE GRSS IDEA initiatives, and participates in the IEEE Young Professionals Affinity Group of the Italy Section.
    \end{IEEEbiography}

    \begin{IEEEbiography}[{\includegraphics[width=1in,height=1.15in,clip,keepaspectratio]{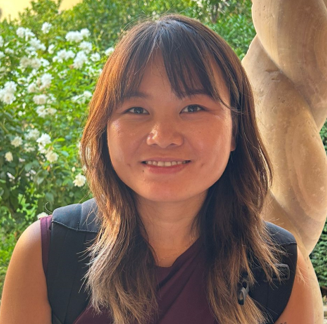}}]{Wenyu Yang}
    (Graduate Student Member, IEEE) was born in Cangzhou, China. She received the B.S. and M.S. degrees in information and communication science from the Beijing Institute of Technology, Beijing, China, in 2017 and 2020, respectively. She is currently pursuing the Ph.D. degree in information and communication technology and engineering with the Department of Engineering, University of Naples Parthenope, Naples, Italy. Her research topics focus on applying deep learning methods to polarimetry synthetic aperture radar (SAR) and SAR tomography images for canopy and terrain topography reconstruction.
    \end{IEEEbiography}

    \begin{IEEEbiography}[{\includegraphics[width=1in,height=1.15in,clip,keepaspectratio]{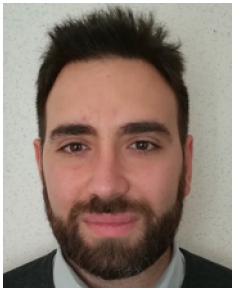}}]{Sergio Vitale}
    (Member, IEEE) received the master’s
    degree (summa cum laude) in telecommunication
    engineering from the University of Naples Federico
    II, Naples, Italy, in May 2017, and the Ph.D. degree
    in information and communication technology and
    engineering from the Department of Engineering,
    University of Naples Parthenope, Naples, in 2021.
    He is currently a Post-Doctoral Researcher at the
    Department of Engineering, University of Naples
    Parthenope. His research activities focus on the
    definition and analysis of Deep Learning solutions for
    image processing applied to Remote Sensing. He mainly focuses on the study and definition of convolutional neural networks (CNNs) for speckle denoising of synthetic aperture radar (SAR) image, SAR tomography, and pansharpening of image from optical sensors. Dr. Vitale won the 2017 Best Italian Remote Sensing Thesis Prize for the IEEE Geoscience Remote Sensing South Italy Chapter in 2018. Since 2021, he has been a member of the Topic Advisor Panel of Sensors (MDPI) and a Guest Editor for the Special Issue of Sensors (MDPI) and the Special Issue of Remote Sensing (MDPI). 
    \end{IEEEbiography}

    \begin{IEEEbiography}[{\includegraphics[width=1in,height=1.15in,clip,keepaspectratio]{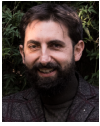}}]{Giampaolo Ferraioli} (Senior Member, IEEE) was born in Lagonegro, Italy, in 1982. He received the B.S. and M.S. degrees and the Ph.D. degree in telecommunication engineering from the Università degli Studi di Napoli Parthenope, Naples, Italy, in 2003, 2005, and 2008, respectively. He has been a Visiting Scientist at the Département TSI, Télécom ParisTech, Paris, France. He is currently an Associate Professor with the Università degli Studi di Napoli Parthenope. His main research interests deal with statistical signal and image processing, radar systems, synthetic aperture radar, image restoration, and magnetic resonance imaging. He serves on the Editorial Board of Remote Sensing (MDPI). In 2009, he won the IEEE 2009 Best European Ph.D. Thesis in Remote Sensing Prize, sponsored by the IEEE Geoscience and Remote Sensing Society. He serves as an Associate Editor for IEEE Geoscience and Remote Sensing Letters and IEEE Journal of Miniaturization for Air and Space Systems.
    \end{IEEEbiography}
    
    \begin{IEEEbiography}[{\includegraphics[width=1in,height=1.15in,clip,keepaspectratio]{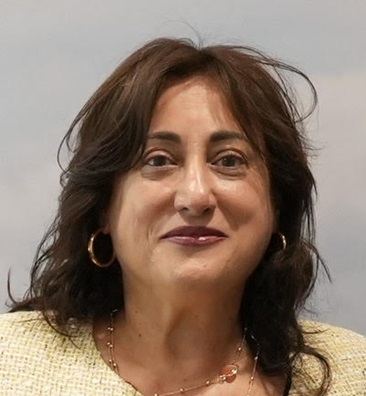}}]{Silvia Liberata Ullo}
    IEEE Senior Member, President of IEEE AESS Italy Chapter, Industry Liaison for IEEE Joint ComSoc/VTS Italy Chapter since 2018, National Referent for FIDAPA BPW Italy Science and Technology Task Force (2019-2021). Member of the Image Analysis and Data Fusion Technical Committee (IADF TC) of the IEEE Geoscience and Remote Sensing Society (GRSS) since 2020. Graduated with laude in 1989 in Electronic Engineering at the University of Naples (Italy), pursued the M.Sc. in Management at MIT (Massachusetts Institute of Technology, USA) in 1992. Researcher and teacher since 2004 at the University of Sannio, Benevento (Italy). Member of Academic Senate and PhD Professors’ Board. Courses: Signal theory and elaboration, Telecommunication networks (Bachelor program); Earth monitoring and mission analysis Lab (Master program), Optical and radar Remote Sensing (Ph.D. program).  Authored 90+ research papers, co-authored many book chapters and served as editor of two books. Associate Editor of relevant journals (IEEE TGRS, JSTARS, GRSL, MDPI Remote Sensing, Springer Arabian Journal of Geosciences and Recent Advances in Computer Science and Communications). Co-Editor-in-Chief of IET Image Processing. Guest Editor of many special issues. Research interests: signal processing, radar systems, sensor networks, smart grids, remote sensing, satellite data analysis, machine learning and quantum ML.
    \end{IEEEbiography}

    \begin{IEEEbiography}[{\includegraphics[width=1in,height=1.15in,clip,keepaspectratio]{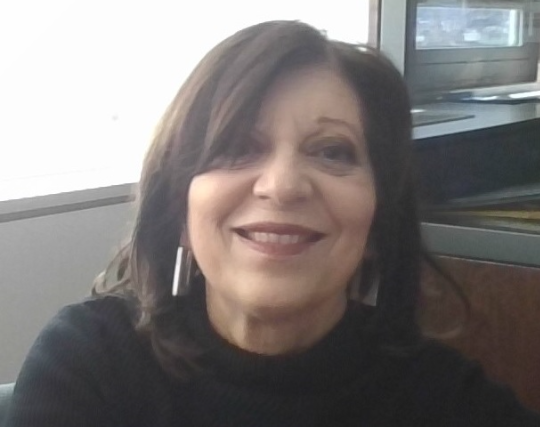}}]{Gilda Schirinzi}
    graduated cum laude in electronic engineering at the University of Naples "Federico II." From 1985 to 1986, she was at the European Space Agency, ESTEC, Noordwijk, The Netherlands. In 1988, she joined the Istituto di Ricerca per l'Elettromagnetismo e i Componenti Elettronici (IRECECNR), Naples, Italy. In 1998, she joined the University of Cassino, Italy, as an associate professor of telecommunications, and in 2005, she became a full professor. Since 2008, she has been at the University of Naples “Parthenope.” Her main scientific interests are in the field of signal processing for Remote Sensing applications, with particular reference to synthetic aperture radar (SAR) interferometry and tomography. She is a Senior Member of the IEEE.
    \end{IEEEbiography}

\end{document}